\documentclass[10pt, journal, compsoc]{IEEEtran}

\usepackage{hyperref}
\usepackage{enumitem}
\usepackage{bbm}
\usepackage{stackengine}
\usepackage[export]{adjustbox}
\usepackage{algorithm}
\usepackage{algorithmic}
\usepackage{ragged2e}
\usepackage{graphicx}
\usepackage{float}
\usepackage{subfigure}
\usepackage{amsmath}
\usepackage{booktabs}
\usepackage{url}
\usepackage{multirow}
\usepackage{colortbl} 
\usepackage{xcolor}
\usepackage{array} 

\ifCLASSOPTIONcompsoc
    \usepackage[nocompress]{cite}
\else
    \usepackage{cite}
\fi

\begin{document}

\title{Computer-aided Colorization State-of-the-science: A Survey}
\author{
        Yu~Cao,
        Xin~Duan,
        Xiangqiao~Meng,
        P.~Y.~Mok,~\IEEEmembership{Member,~IEEE},\\
        Ping~Li,~\IEEEmembership{Member,~IEEE},
        and Tong-Yee~Lee,~\IEEEmembership{Senior Member,~IEEE}

\IEEEcompsocitemizethanks{
\IEEEcompsocthanksitem Yu~Cao is with the School of Fashion and Textiles, The Hong Kong Polytechnic University, Hong Kong. E-mail: yu-daniel.cao@connect.polyu.hk.
\IEEEcompsocthanksitem Xin~Duan, Xiangqiao~Meng, and Ping~Li are with the Department of Computing and the School of Design, The Hong Kong Polytechnic University, Hong Kong. E-mail: \{hizuka.duan, xiangqiao.meng\}@connect.polyu.hk, p.li@polyu.edu.hk.
\IEEEcompsocthanksitem P.~Y.~Mok is with the School of Fashion and Textiles, The Hong Kong Polytechnic University, Hong Kong, and the Laboratory for Artificial Intelligence in Design, Hong Kong. E-mail: tracy.mok@polyu.edu.hk.
\IEEEcompsocthanksitem Tong-Yee Lee is with the Department of Computer Science and Information Engineering, National Cheng-Kung University, Tainan 70101, Taiwan.\protect\\
E-mail: tonylee@ncku.edu.tw.
}
\thanks{Yu~Cao, Xin~Duan contributed equally to this work.}
\thanks{P.~Y.~Mok, Ping~Li, and Tong-Yee Lee are corresponding authors.}
}


\IEEEtitleabstractindextext{
\begin{abstract}
\justifying
This paper reviews published research in the field of computer-aided colorization technology. We argue that the colorization task originates from computer graphics, prospers by introducing computer vision, and tends to the fusion of vision and graphics, so we put forward our taxonomy and organize the whole paper chronologically.
We extend the existing reconstruction-based colorization evaluation techniques, considering that aesthetic assessment of colored images should be introduced to ensure that colorization satisfies human visual-related requirements and emotions more closely. We perform the colorization aesthetic assessment on seven representative unconditional colorization models and discuss the difference between our assessment and the existing reconstruction-based metrics.
Finally, this paper identifies unresolved issues and proposes fruitful areas for future research and development.
Access to the project associated with this survey can be obtained at \href{https://github.com/DanielCho-HK/Colorization}{https://github.com/DanielCho-HK/Colorization}.
\end{abstract}

\begin{IEEEkeywords}
Colorization, computer graphics, computer vision, colorization aesthetic assessment.
\end{IEEEkeywords}}

\maketitle

\IEEEdisplaynontitleabstractindextext

\IEEEpeerreviewmaketitle

\IEEEraisesectionheading{\section{Introduction}\label{sec:introduction}}
\IEEEPARstart{C}{olor} is an integrated and crucial part of the real world.
While appreciating the world's natural beauty, humans have never stopped trying to capture the rich colors of nature by utilizing methods from painting to photography.
A Canadian, Wilson Markle, first invented computer-aided colorization technology in 1970 to add color to monochrome footage of the moon obtained during the Apollo program. Nowadays, such colorization technologies have a wide range of applications, including restoring the original colors for black-and-white photos~\cite{xu23}, legacy 
films~\cite{iizuka2019deepremaster}, cartoon~\cite{varga2017automatic,li21cartoon,xing2024tooncrafter}, animation colorization~\cite{dai2024learning,huang2024lvcd}, etc.
Color is also an indispensable element of the digital world, such as in computer graphics research~\cite{truckenbrod81}, a significant component of computer-aided visualization of information, concepts, and ideas. 

Colorization can be defined as the generation of color information from gray-scale images and line drawings (or sketches) while keeping the structural details unchanged. For more than two decades, this research has attracted the attention of many researchers in computer graphics and computer vision to tackle three main problems: (1) the ill-posed nature of recovering diverse color information~\cite{charpiat2008automatic} while performing gray-scale image or video colorization; (2) consider semantic understanding and different color sources~\cite{yin21} into colorization tasks; (3) extend the colorization target from gray-scale image to other formats like manga, sketch, and line drawing.

To generate various colored results, the existing colorization techniques can be roughly divided into user-guided information, explicitly performing diverse colorization, and color prior learned implicitly from large-scale datasets that represent different color sources. 
Besides, the `semantic’ refers to fine-grained object recognition and structure region differences in natural image and line art colorization scenarios.
Especially in reference-based and text-based colorization tasks, this cross-domain semantic understanding ability mainly relies on deep learning techniques. Similarly, in the line drawing colorization, researchers proposed a region-segmented method that explicitly distinguishes different semantic regions, which can alleviate color bleeding problems. In the existing coloring literature, more and more articles are designed for manga, line drawing, and sketches in addition to processing gray-scale images. Chen~et al.~\cite{CHEN202251} gives a detailed comparison of these non-photorealistic colorization targets, and in our survey, we call them line drawing generally. Unlike gray-scale images, line art encounters more challenges, including information sparsity, lack of high-quality paired data for training, complicated line structures with unique tones and textures in anime and manga, etc. Hence, specific-designed methods for line drawing colorization often utilize user hints or reference images as conditionals.

\begin{figure*}[h]
    \centering
    \includegraphics[width=1.0\linewidth]{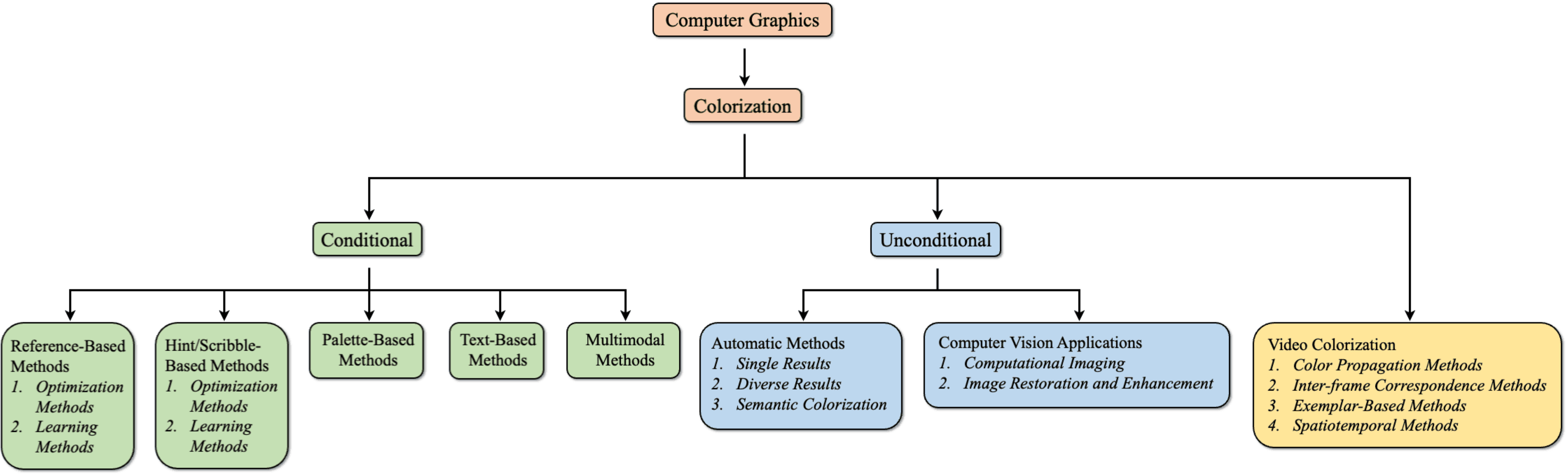}
    \vspace{-10px}
    \caption{Taxonomy of colorization technology. We classify colorization as a sub-task of computer graphics, in which the green part belongs to the explicit diverse colorization using conditional controls, and the blue part belongs to the colorization methods and applications using computer vision technology. We individually mark the video colorization in yellow to represent it as an extension of image colorization in spatiotemporal dimensions.}
    \label{fig:taxonomy}
\end{figure*}

Although some excellent reviews have been published on colorization, they are not considered sufficiently comprehensive. For example, Anwar~et al.~\cite{anwar2020image} mainly focused on single-image colorization, while Huang~et al.~\cite{HUANG2022105006} focused on deep learning-based colorization and largely ignored traditional coloring methods. On the other hand, Chen~et al.~\cite{CHEN202251} reviewed colorization from image analogy to learning-based methods but did not discuss the latest advancements, and they proposed their taxonomy according to various types of colorization targets.
Our survey provides a much more detailed, comprehensive, and up-to-date review of computer-aided colorization. We present our taxonomy from the methodological perspective of existing literature. Considering colorization as a kind of computer graphics task, we divide colorization into three broad categories: conditional methods, unconditional methods, and video colorization (see Fig.~\ref{fig:taxonomy}).
It complements the previous reviews, thereby completing the analysis of the development of coloring technologies, and also identifies potentially rewarding future research directions in coloring technology. Given that existing reconstruction-based evaluation metrics are not initially designed for colorization tasks and there is no ground truth image when performing colorization. The quality of image coloring is an abstract concept that cannot be easily quantified, such as emotion and aesthetics.
Inspired by the existing CLIP-based image quality assessment research, we propose a novel colorization aesthetic assessment method to simulate a human vision perception system.
The main contributions of the paper can be summarized as follows:
\begin{itemize}[itemsep=0pt, topsep=0pt, parsep=0pt]
    \item We thoroughly review the research and other materials on colorization technology published during the past two decades, providing a precise, insightful literature-based analysis for follow-up research.
    \item We introduce image aesthetic assessment into colorization tasks and evaluate seven unconditional image colorization methods, this being the first time this is done to our knowledge.
    \item We discuss some challenging issues and make suggestions concerning the development trends and potentially fruitful future research directions and technological progress.
\end{itemize}

The remainder of this survey is structured as follows.
Section~\ref{sec:colorization} reviews the relevant literature about colorization technology organized according to our proposed taxonomy. Section~\ref{sec:dataset} covers representative datasets for training learning-based colorization models. 
Section~\ref{sec:metric} discusses the concept of colorization aesthetic assessment and compares seven unconditional colorization techniques with the proposed new aesthetic assessment. 
Section~\ref{sec:discussion} discusses future research directions, and Section~\ref{sec:summary} contains a summary and conclusions.

\section{Colorization}\label{sec:colorization}

\subsection{Conditional Colorization}
Conditional colorization is a kind of colorization technology that can explicitly generate diverse colored results according to different types of user controls. Existing conditional methods consist of five main controls: reference image, hint/scribble, palette, text, and multi-modal controls. The earliest natural image coloring articles we have traced are based on reference images~\cite{Welsh02} and user hints~\cite{levin04}. With the progress of technology, the coloring method has also evolved from the traditional non-parametric optimization paradigm to the learning-based paradigm. The colorization targets have also extended from gray-scale images to line drawings, including manga~\cite{qu06,fu17}, anime~\cite{cao23,yan23,animediffusion24,lin24}, cartoons~\cite{LIU201878,liu2022reference,chen22}, icons~\cite{sun19,li22}, etc.

\subsubsection{Reference-based Methods}
\begin{figure}[ht]
    \centering
    \includegraphics[width=1.0\linewidth]{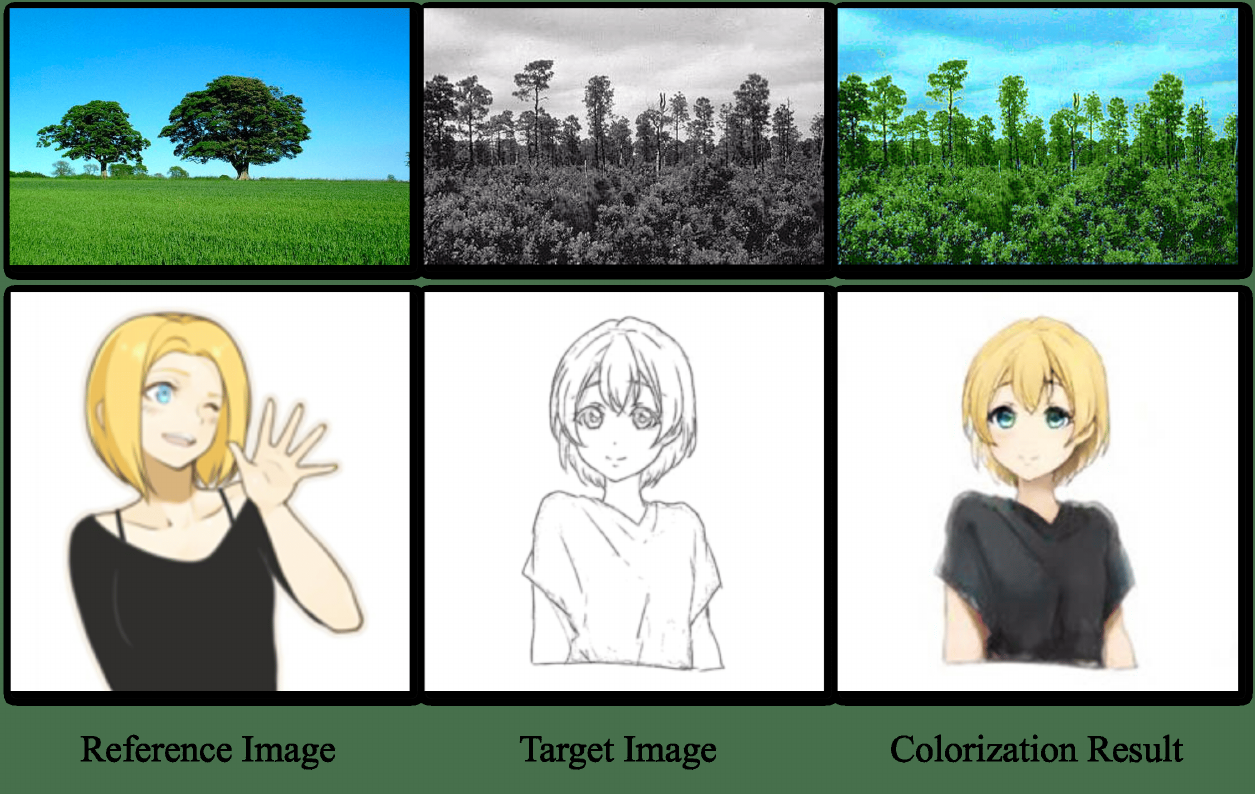}
    \vspace{-10px}
    \caption{Illustration of reference-based colorization. The top row shows reference-based gray-scale image colorization and the bottom row shows reference-based line drawing colorization. Images courtesy of~\cite{Welsh02,cao23}.}
    \label{fig:reference_colorization}
\end{figure}

Reference-based colorization is a method that converts color information from reference images to target gray-scale images or black-and-white line drawings (see Fig.~\ref{fig:reference_colorization}).

{\flushleft \bf Gray-scale Image Colorization.}
Reference-based gray-scale image colorization originates from the concept of image analogy~\cite{analogy}, a method of automatically learning an image filter from training data. As in the case of other classic CG tasks, such as texture synthesis, texture transfer, and artistic rendering, reference-based image colorization can be considered an image filter simulation using image analogy.
Inspired by this, Welsh~et al.~\cite{Welsh02} proposed the first color-transferring method from a source color image to a target gray-scale image. Their basic idea was first to perform pixel neighborhood matching in the luminance channel and then transfer chromaticity values from the source to the target. 
Irony~et al.~\cite{Irony05} introduced an exemplar-based colorization technique incorporating a specially designed texture-based classifier for more accurate localized transfer. This classifier is derived from a detailed analysis of low-level features in the reference image.
In the early work, gray-scale image colorization models were built upon the assumption that similarities in gray-scale intensities indicate color similarities.
Colorization algorithms can be misled by intensity disparities arising from variations in shades and brightness between its reference image and the target image.
To address the problem of illumination inconsistency between target and reference images, Liu~et al.~\cite{liu-2008-intrinsic} introduced an intrinsic colorization approach that involved computing an illumination-independent reference image through intrinsic image decomposition. To mitigate the problems caused by variations in illumination, their method leveraged multiple reference images obtained from the internet.
A similar concept of using an extensive online image database to colorize gray-scale images was also proposed by Morimoto~et al.~\cite{morimoto09}.
In addition, the success of such reference-based methods depends heavily on retrieving a suitable reference image. To expand the colorization targets ranging from famous landmarks to general objects and scenes, Chia~et al.~\cite{chia11} proposed a more user-friendly colorization method that utilized semantic labels to search appropriate reference photos from the internet and facilitate accurate color transfer.
To improve the speed of image colorization, Gupta~et al.~\cite{gupta2012image} proposed a method that used a fast cascade feature matching scheme to rapidly find correspondence between reference images and target images at the superpixel level.
Bugeau~et al.~\cite{bugeau2013variational} designed a variational model to simultaneously model color selection and spatial constraints.
Li~et al.~\cite{li17} presented a novel approach for colorizing target superpixels by formulating it as a problem of dictionary-based sparse reconstruction. They introduced the first sparse pursuit image colorization method, using a single reference image.
Reducing the limitations of images provided by users and improving the precision of dense feature matching between images can make reference-based colorization technology more versatile.
Generally, it is challenging when users provide reference and target images with objects with different scales.
To tackle the problem of feature matching in different scales, 
Li~et al.~\cite{li19} introduced a cross-scale matching technique incorporating localized considerations of various potential scales during the matching process. They subsequently employed the graph-cut technique to fuse the scales globally, aiming to identify spatially coherent scales that exhibit high-quality matching.
Fang~et al.~\cite{fang20} approached exemplar-based image colorization as a problem of color selection, incorporating regularization constraints. They focused on utilizing superpixels as processing units to enhance both the efficiency and robustness of colorization. Notably, their work introduced the utilization of superpixel-based non-localized self-similarity and localized spatial consistency as novel techniques for image colorization.

\textit{The technological evolution route of traditional reference-based gray-scale image colorization is inspired by single image analogy, considering multiple reference images, then utilizing the retrieval-based method for selecting appropriate reference images until the superpixel became the most mature solution among the traditional techniques.}

Since 2012, with the successful applications of Deep Convolutional Neural Networks (CNN) model~\cite{krizhevsky2012imagenet} for many vision-related studies, reference-based image colorization methods have also changed from traditional optimization-based non-parametric models to CNN-based parametric models. 
To address the complex scenario, where clear correspondences between the target and the reference images are lacking, He~et al.~\cite{he18} proposed an end-to-end network to learn the selection, propagation, and prediction of colors from existing data. Unlike earlier deep learning-based colorization approaches, their method effectively captured local and global contents, resulting in improved colorization outcomes, which avoided semantic inaccuracies or dullness.
Xiao~et al.~\cite{xiao2020example} presented a Dense Encoding Pyramids Network for colorization, which mapped the color distribution from a reference image onto a gray-scale image. Their model incorporated novel Parallel Residual Dense Blocks to capture comprehensive local-global context information. Furthermore, a Hierarchical Decoder-Encoder Filter was employed to aggregate color distribution results across adjacent feature maps.
Inspired by photorealistic image stylization work~\cite{huang2017arbitrary,li2018closed}, Xu~et al.~\cite{xu20} designed a deep learning model consisting of a transfer net and a colorization net to perform real-time exemplar-based image colorization.
For a reliable reference-based image colorization system, the semantic colors associated with the objects and global color distributions are important color characteristics of the reference image. 
Lu~et al.~\cite{lu20} introduced Gray2ColorNet to address this aspect. This end-to-end deep neural network innovatively combined reference images' semantic and global colors for effective image colorization.
Yin~et al.~\cite{yin21} modeled colorization as a query-assignment problem for different color sources and designed a unified attention mechanism-based framework. Under this unified framework, selecting and assigning colors from the reference images to the gray-scale image adhered to a shared criterion, utilizing the semantic features as a primary factor.
Inspired by the observation that a broad learning system was capable of efficiently extracting semantic features~\cite{chen18}, 
Li~et al.~\cite{li21} proposed Broad-GAN, an approach for semantic-aware image colorization. They devised a customized loss function to enhance training stability and evaluate the semantic similarity between the target and ground-truth images.
Attention-aware methods have recently emerged to tackle the inherent semantic correspondence problem in reference-based image colorization. 
Carrillo~et al.~\cite{Carrillo_2022_ACCV} introduced a super-attention block that leveraged superpixel features to transfer semantically related color characteristics from a reference image across various scales within a deep learning network. Similarly, Bai~et al.~\cite{bai2022semantic} proposed a Semantic-Sparse Colorization Network that employed a sparse attention mechanism to transfer global image style and detailed semantic-related colors to gray-scale images in a coarse-to-fine manner. Wang~et al.~\cite{wang23} developed an effective exemplar colorization strategy utilizing a pyramid dual non-local attention network to explore long-range dependencies and multi-scale correlations.

\textit{Learning-based reference colorization research mainly tackled the semantic correspondence problem by introducing different solutions, including local-global deep features aggregation, exemplar-based style transfer formulation, and attention mechanism modules. In addition, decoupling the color sources when performing the reference-based colorization is significant. Usually, the color comes from 1) the semantic colors linked to objects in the reference image, 2) the global color distribution encompassing tones and hues of the reference image, and 3) the color information learned from large-scale datasets.}

{\flushleft \bf Line Drawing Colorization.} 
The early method~\cite{zhang17} formulated reference-based line drawing colorization as a neural style transfer problem~\cite{gatys16}. Since naive neural style transfer methods designed for natural images cannot deal with line drawings, they incorporated a residual U-Net architecture and utilized an Auxiliary Classifier Generative Adversarial Network (AC-GAN)~\cite{pmlr-v70-odena17a} to apply the desired style to the gray-scale sketch.
In line drawing colorization, a specialized group of work focuses on `manga' and `anime,' namely comics and animations originating in Japan. Furusawa~et al.~\cite{fu17} introduced Comicolorization as the pioneering approach to colorizing complete manga titles, encompassing sets of manga pages.
The semi-automatic system uses reference images to colorize the input manga images.
In the anime creation industry, artists often manually draw anime character illustrations with empty pupils first and then fill in the preferred colors or details in the pupils.
Akita~et al.~\cite{akita2020colorization} introduced a colorization model combined with a pupil position estimation module to colorize anime character faces automatically with accurate pupil colors.
To solve the problem of not being able to obtain paired training data before and after colorization, Lee~et al.~\cite{lee20} proposed a training scheme aimed at learning visual correspondence. They achieved this by generating self-augmented references with a self-supervised training scheme, eliminating the need for manually annotated labels of visual correspondence. This development facilitated end-to-end network optimization without explicit supervision.
Cao~et al.~\cite{cao2021line} developed a segmentation fusion model to mitigate the problem of color-bleeding artifacts effectively.
Li~et al.~\cite{li2022eliminating} identified the gradient conflict in attention modules during line-art colorization, where a gradient branch would exhibit negative cosine similarity with the aggregated gradient, negatively impacting the training stability. To address this issue, they proposed a training strategy called Stop-Gradient Attention. This strategy eliminated the gradient conflict problem, enabling the model to learn improved colorization correspondence.
Liu~et al.~\cite{liu2022reference} employed a multi-scale discriminator to enhance the visual realism of colorized cartoons, focusing on improving both global color composition and local color shading. 
Chen~et al.~\cite{chen22} introduced an active learning-based framework that combined local region matching between line art and reference-colored images, followed by spatial context refinement using mixed-integer quadratic programming (MIQP).
Cao~et al.~\cite{cao23} devised an explicitly attention-aware model for generating high-quality colored anime line drawing images. 
Wu~et al.~\cite{wu23} proposed SDL, a pioneering self-driven dual-path framework for limited data reference-based line art colorization. 
More recently, Cao~et al.~\cite{animediffusion24} introduced AnimeDiffusion, the first diffusion model tailored explicitly for reference-based colorization of anime face line drawings.

\textit{Reference-based line drawing colorization research focused on cross-domain semantic correspondence and color consistency to generate appropriate colored results. Due to the lack of pairs of high-quality training data, it is tough to train models in a supervised manner. Since there is a lack of large-scale training data like natural images, the generalization of existing models is still limited.}

\subsubsection{Hint/Scribble-based Methods}
\begin{figure}[ht]
    \centering
    \includegraphics[width=1.0\linewidth]{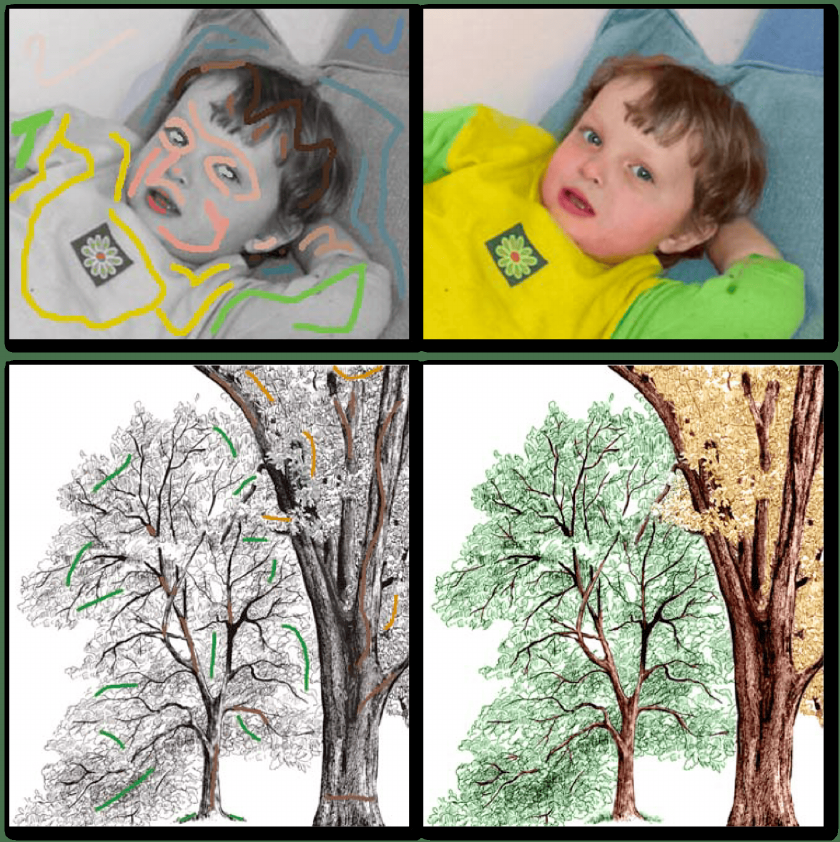}
    \vspace{-10px}
    \caption{Illustration of colorization using color scribbles (or strokes). The top row shows gray-scale image colorization, and the bottom row shows the colorization of line drawings. Images courtesy of~\cite{levin04,qu06}.}
    \label{fig:color_scribbles}
\end{figure}

\begin{figure}[ht]
    \centering
    \includegraphics[width=1.0\linewidth]{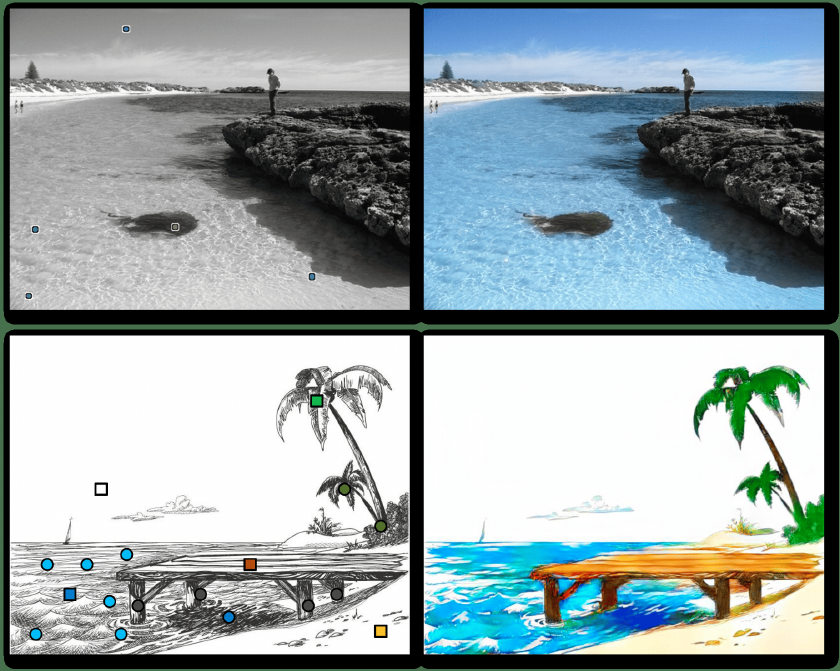}
    \vspace{-10px}
    \caption{Illustration of colorization using color points. The top row shows gray-scale image colorization, and the bottom row shows the colorization of the line drawing. Images courtesy of~\cite{zhang17,zhang18}.}
    \label{fig:color_points}
\end{figure}

Hint/Scribble-based colorization is a technology that performs colorization by propagating local user-provided color scribbles (Fig.~\ref{fig:color_scribbles}) and color points (Fig.~\ref{fig:color_points}). 

{\flushleft \bf Gray-scale Image Colorization.}
Levin~et al.~\cite{levin04} introduced an innovative interactive colorization technique that assumed neighboring pixels in space-time, with similar intensities, should exhibit similar colors.
Using a quadratic cost function, they formulated the color propagation process for an optimization problem.
Manual scribbling can be tedious for images with complex details and requires aesthetic skills to obtain realistic results. The method developed by Irony~et al.~\cite{Irony05} can automatically generate `micro-scribbles' from the image the user provides as an example, greatly facilitating user operation.
To solve the `color bleeding' problem in the boundary regions, Huang~et al.~\cite{huang05} developed an adaptive edge detection scheme to prevent colorization from bleeding over boundaries.
Yatziv and Sapiro~\cite{ys06} designed a fast colorization framework based on the concept of color blending to speed up the colorization process, and their method can be easily extended to video colorization without the need for optical flow computation.
Since previous user-guided methods require a vast number of user inputs (e.g., in the form of strokes) to achieve high-quality colorization of images with complex textures, Luan~et al.~\cite{luan07} considered the colorization problem as one of image segmentation by using texture cluster, and proposed an easy two-step colorization system including Color Labeling and Color Mapping.

After 2017, user-guided colorization methods have entered the era of deep learning.
Zhang~et al.~\cite{richard17} introduced deep learning technology to assist users in making decisions of informed color inputs. Through training their model on large-scale natural image datasets, the model acquires fundamental abilities for image semantic recognition and color information statistics. Combined with the developed interactive system, users can specify their preferences to generate plausible colorization results.
Kim~et al.~\cite{kim21} designed a simple add-on edge-enhancing network to allow users to annotate the color bleeding region with scribbles interactively. Instead of performing a direct operation on the image level, their model used scribbles and intermediate feature maps to generate edge-enhanced colorization outputs.
Xiao~et al.~\cite{xiao22} designed a two-stage deep colorization model that simultaneously supported local color points and global palette inputs.
Yun~et al.~\cite{yun23} proposed the first color point-based real-time colorization model based on the Vision Transformer~\cite{dosovitskiy2020vit}. Their method, leveraging the self-attention mechanism, can avoid producing partially colorized results. 

{\flushleft \bf Line Drawing Colorization.}
User-guided colorization methods for line drawings can be traced back to the work of~\cite{qu06}, who developed a stroke-based approach tailored explicitly for colorizing manga, characterized by intensive strokes, hatching, halftoning, and screening. Due to the rough continuity of patterns in the manga, conventional intensity-continuity techniques are ineffective. They proposed a method to propagate the color in pattern-continuous and intensity-continuous modes automatically.
Sýkora~et al.~\cite{sykora2009lazybrush} presented a flexible colorization tool that could be applied to various drawing styles instead of the previous style-limited approaches.

As in the case of natural images, starting in 2017, many user-guided line art colorization methods using deep learning techniques were proposed. Sangkloy~et al.~\cite{sangkloy2017scribbler} formulated the colorization process as a sketch-to-image synthesis problem with scribbles control. They were the first to utilize GAN to generate realistic images according to sketches and sparse color scribbles.
Liu~et al.~\cite{LIU201878} designed the auto-painter model, based on conditional Generative Adversarial Networks (cGAN)~\cite{mirza2014conditional}, which automatically generates colorized cartoon images from a sketch.
Ci~et al.~\cite{ci18} proposed a model based on cGAN architecture for scribble-based anime line art colorization. To generate authentic illustrations with accurate shading, they integrated their framework with WGAN-GP~\cite{wgan-gp} and perceptual loss~\cite{johnson2016perceptual}.
Zhang~et al.~\cite{zhang18} proposed a learning-based framework consisting of two distinct stages: drafting and refinement.
This decomposition simplified the learning process and enhanced the overall quality of the final colorization outcomes.
In line art colorization, flat filling is a process that uses relatively flat colors rather than color textures. 
Zhang~et al.~\cite{zhang2021user} proposed the Split Filling Mechanism framework to control the influence areas of user scribbles and generate realistic color combinations. Their method can fill in flat, consistent colors to regions instead of pixel-level color textures.
Yuan and Simo-Serra~\cite{yuan2021line} presented a Concatenation and Spatial Attention module that can generate consistent and high-quality line art colorization from user inputs.
Previous methods performed colorization in RGB color space, which resulted in dull colors and inappropriate saturations. 
To address this problem, Dou~et al.~\cite{dou21} introduced the DCSGAN model, which was the first to utilize Hue, Saturation, and Value (HSV) color space to enhance anime sketch colorization. The HSV color space closely aligns with the human visual cognition system. It is well-suited for colorization tasks incorporating prior human drawings, including hue variation, saturation contrast, and gray contrast.
Recently, Carrillo~et al.~\cite{carrillo2023diffusart} introduced an interactive method for colorizing line art using conditional Diffusion Probabilistic Models.
Cho~et al.~\cite{cho23} developed the GuidingPainter model to improve the efficiency of the interactive sketch colorization process, based on the concept that making the model actively seeks regions where color hints would be provided rather than rely too heavily on deciding color local-position information by users.

\textit{Hint/scribble-based methods solved the color bleeding problems by adding restrictions on the boundaries of different semantic regions. Researchers also tried to reduce the color hint inputs to ease the user's effort to perform coloring.}

\subsubsection{Palette-based Methods}
\begin{figure}[t]
    \centering
    \includegraphics[width=1.0\linewidth]{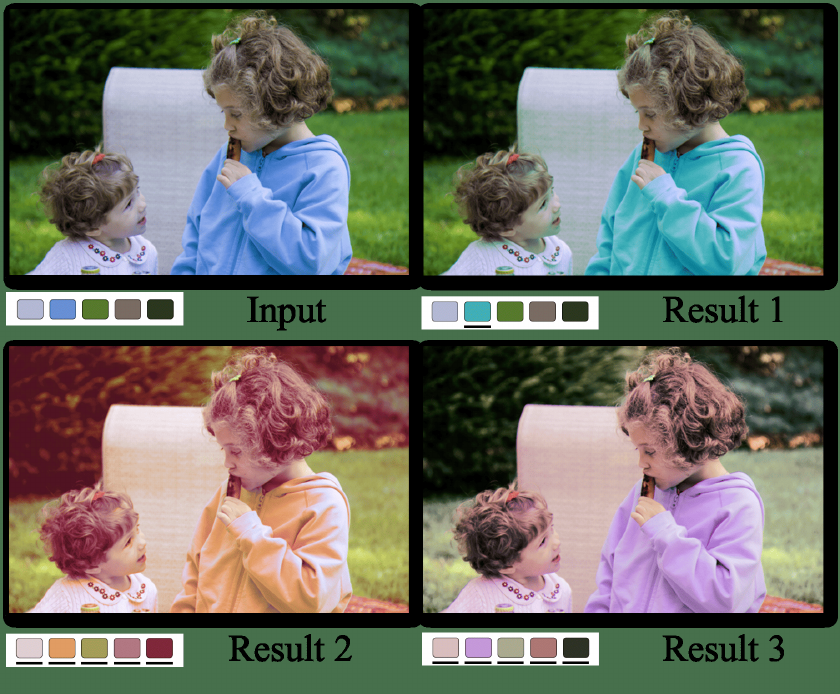}
    \vspace{-10px}
    \caption{Palette-based colorization method is suitable for image recolorization applications. Users can directly change one or all color elements of the palette extracted from the input image. The black line below the color element indicates the color users select to edit. Images courtesy of~\cite{chang2015palette}.}
    \label{fig:palette}
\end{figure}

Palette-based colorization is a method that allows users to perform global color edits by manipulating a small set of representative colors, as illustrated in Fig.~\ref{fig:palette}. 
The color palette can usually reflect the overall tones or themes of the image, and image color conveys emotions through color themes. 
Wang~et al.~\cite{wang2012affective} proposed the first system, called the Affective Colorization System, which can efficiently colorize a gray-scale image semantically using a color palette incorporated with emotional information. 
Bahng~et al.~\cite{bahng2018coloring} regarded the palette as an intermediate representation that conveys the semantics of the image. They first designed a model to generate multiple palettes according to different text inputs and then performed palette-based gray-scale image colorization.
Xiao~et al.~\cite{xiao2020example} proposed a reference-based colorization framework, supporting a palette as one reference type to generate realistic colored results.
Wu~et al.~\cite{wu2023} introduced a flexible icon colorization model based on guided images and palettes.

\textit{The goal of palette-based image color transfer is to apply the color characteristics of a predefined color palette to a target image. Because only a limited number of global color palettes can be modified, the image is prone to color bleeding or the overall tone disharmony problem, as illustrated in Fig.~\ref{fig:palette}.
Compared to other conditional colorization technologies, palette-based methods attracted relatively less attention from researchers. Publications dealing with palette-based methods mainly cover natural gray-scale image colorization or recolorization~\cite{chang2015palette,cho17,du2024palette}.}

\subsubsection{Text-Based Methods}
\begin{figure}[ht]
    \centering
    \includegraphics[width=1.0\linewidth]{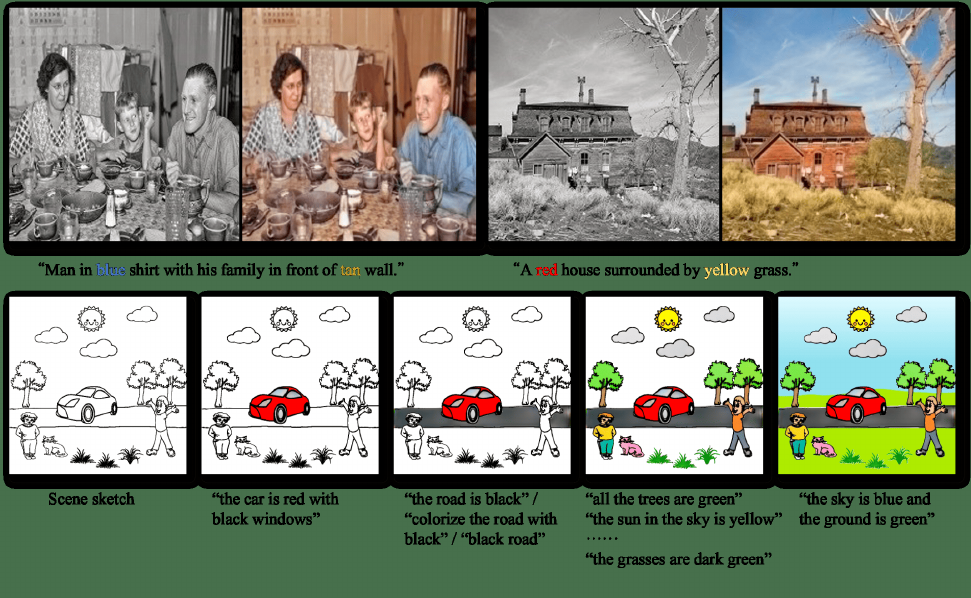}
    \vspace{-10px}
    \caption{Illustrations of text-based colorization. The top row illustrates gray-scale image colorization and the bottom row line drawing colorization. Images courtesy of~\cite{chang2022coder,zou2019language}.}
    \label{fig:text}
\end{figure}

Text-based colorization~\cite{manjunatha2018learning} performs coloring based on users' guidance in natural language instructions, and it is a relatively novel cross-modal interactive method both in natural and line images, as illustrated in Fig.~\ref{fig:text}.

Chen~et al.~\cite{chen2018language} pioneered a text-based method for colorization.
They developed a comprehensive modeling framework that completes two intertwined tasks of image editing based on language: image segmentation and colorization. Their approach involved using a CNN to process the source image and an LSTM~\cite{lstm} network to encode the textual features representing the language expression.
Another vital approach was presented by Bahng~et al.~\cite{bahng2018coloring}, which focused on linking specific words with particular colors to encapsulate the semantics of the text input. Their method produced relevant color palettes that captured the essence of the text. The palettes were then applied to add color to a gray-scale input image. They announced their technique allows individuals without artistic expertise to create color palettes that effectively communicate high-level concepts.
Kim~et al.~\cite{kim2019tag2pix} proposed an alternative approach employing a GAN for colorizing line art. Their method used monochromatic line drawings and color tag data as input to generate high-quality colored images. It is worth noting that their tag-based coloring method is the prototype of the current popular ControlNet-based~\cite{zhang2023adding} anime content generation using line drawing guidance, thanks to the Danbooru dataset with detailed tag labeling meta information. 

Weng~et al.~\cite{weng2022code} presented a method that decoupled colors and objects into different spaces. Their approach allowed for correctly applying designated (potentially unusual) color words to objects, successfully addressing the problems associated with the coupling of color and objects and mismatches between them.
Chang~et al.~\cite{chang2022coder} introduced the first transformer-based text-based colorization system by analyzing the cross-modal relationship between images and texts using cross-attention learning, thereby addressing the large gap between the two modalities. 
Early methods independently leveraged two distinct architectures for feature extraction: CNN for images and LSTM for relevant textual captions.
In another study, Chang~et al.~\cite{chang2023coins} proposed a Transformer-based framework to automatically consolidate related image patches and attain instance awareness without additional information.
Their method solved the challenge encountered by previous research in differentiating instances referring to the same object.
Chang~et al.~\cite{chang2023cad} utilized the robust language understanding and extensive color priors provided by Stable Diffusion~\cite{rombach2022high} to enable text-based colorization. Unlike previous approaches, which relied on comprehensive color descriptions for many objects in the image, this method avoided suboptimal performance, especially for items without color-matched descriptive words.

\textit{Text-based colorization is a sub-task of prompt-based content generation. It leverages natural language processing (NLP) and computer vision techniques to interpret and apply colors as described in the text.}

\subsubsection{Multi-modal Methods}
Multi-modal colorization performs colorization by combining different types of control. 
Huang~et al.~\cite{huang2022unicolor} presented UniColor. This first unified framework enables colorization in multiple modalities, encompassing both unconditional and conditional approaches and accommodating various conditions, including stroke, exemplar, text, and combinations. The framework was a two-stage colorization process that integrated these conditions into a single model.
In the initial stage, the diverse multi-modal conditions are transformed into a shared representation, known as hint points. Notably, they introduced a CLIP-based~\cite{radford2021learning} approach to convert textual inputs into hint points, ensuring compatibility with other modalities. The subsequent stage involved a Transformer-based network consisting of Chroma-VQGAN and Hybrid-Transformer components. 
Recently, two methods of diffusion-based natural image colorization~\cite{liang2024control} and anime sketch colorization~\cite{yan23,yan2024colorizediffusion} have been proposed. Both support multiple types of control for interactive colorization.

\textit{With the great improvement of the performance of generative models, multi-modal interactive coloring technology has gradually become the research trend in academics and has high application value.}

\subsection{Unconditional Colorization}
Unconditional image colorization is a process in which a gray-scale image is converted into a color image without additional input or guidance. In past studies, researchers have also referred to it as automatic colorization. The model can learn more abundant prior color information by training on large-scale natural image datasets to generate plausible results. Moreover, thanks to the progress of generative model technology, automatic coloring technology has evolved from generating a random single result to diverse colored results.

\subsubsection{Automatic Methods}
{\flushleft \bf Single Results.}
{\flushleft 1. Regression loss} \\
The early attempts~\cite{cheng2015deep,deshpande2015learning} leveraged neural networks as \textbf{regression models} to minimize average reconstruction errors. The colorization task enters the era of deep learning by using a neural model to learn the mapping between gray-scale images and colored images.
{\flushleft 2. Classification loss} \\
By introducing the concept of representation learning~\cite{bengio13}, Zhang~et al.~\cite{zhang2016colorful} treated colorization as a \textbf{classification task} by using the image's $L$ channel as input and its $ab$ channels as outputs for supervised learning. Changing pixel regression to pixel classification significantly improved the colorization quality, especially regarding saturation.
Similarly, Larsson~et al.~\cite{larsson2016learning,larsson2017colorization} explored colorization as a means of self-supervised learning for visual comprehension, free from the constraints of pre-trained backbone models.
These insights into representation learning through colorization have significantly advanced the understanding of self-supervised tasks and their applications in computer vision.
{\flushleft 3. Image-to-image translation} \\
Colorization can also be formulated as an image-to-image translation task. Santhanam~et al.~\cite{santhanam2017generalized} developed a multi-context image representation that treated image-to-image translation as a regression task, providing a broader localized-context spectrum compared to previous methods. In contrast, cGAN learns a `structured' loss. Isola~et al.~\cite{isola2017image} demonstrated the effectiveness of image-conditional GAN as a general-purpose solution for image-to-image translation by incorporating a discriminator. The model effectively captured intricate texture and style details.

\begin{figure}[h]
    \centering
    \includegraphics[width=1.0\linewidth]{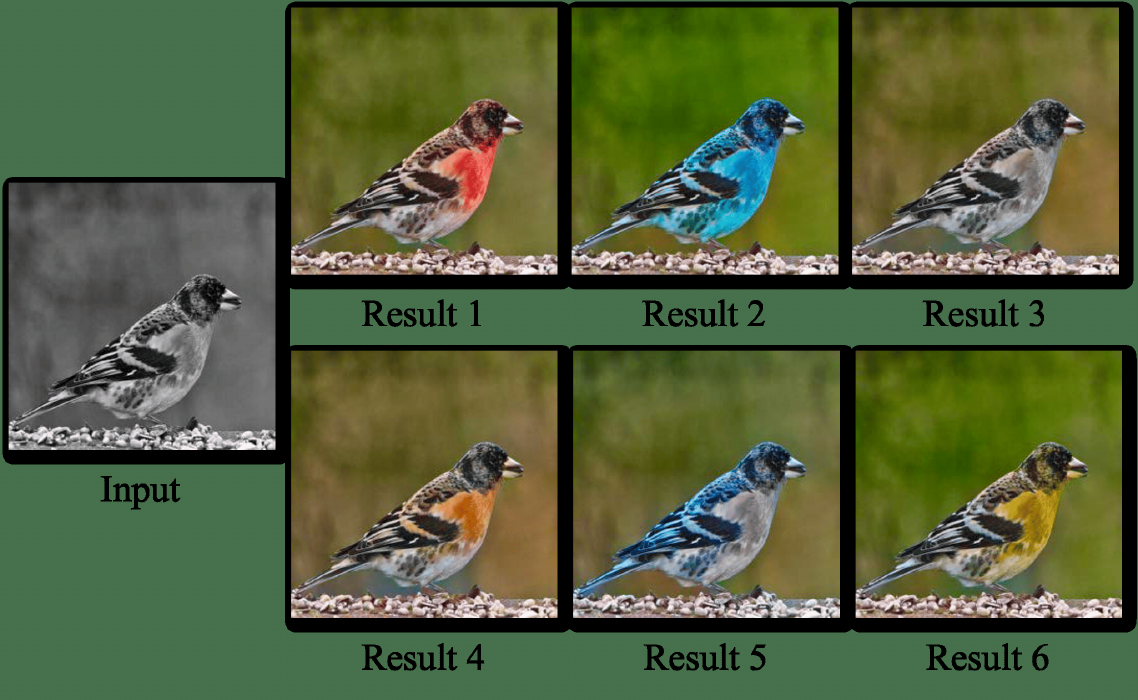}
    \vspace{-10px}
    \caption{Illustration of diverse colorization methods. The model, trained on a large-scale image dataset, automatically learns the color priors and performs implicit diverse colorization. The gray-scale input image is on the left, and the six images on the right are colored results of certain conditioned colorization methods. Images courtesy of~\cite{wu21}.}
    \label{fig:diverse}
\end{figure}

{\flushleft \bf Diverse Results.}
Different from conditional colorization methods involving various types of user control to generate diverse colored results explicitly, by utilizing generative models, automatic methods can perform diverse coloration implicitly, as illustrated in Fig.~\ref{fig:diverse}.
Charpiat~et al.~\cite{charpiat2008automatic} undertook a pioneering work involving implicitly learning the multi-modal probability distribution of colors.
It is worth noting that the word `multi-modality' used in their paper about fifteen years ago meant `diversity of colors,' which differs from the meaning of multi-modality today, namely, multiple forms of sensory input, including images, text, audio, and video. Nevertheless, as a result of the groundbreaking work, many subsequent studies dealing with diverse colorization still used `multi-modal' within the context of `diversity of colors.'

We find that previous studies aimed at diverse colorization published in different periods adopted the contemporary famous generative model algorithms as the backbone of their approaches, including VAE~\cite{Deshpande17,messaoud18}, Flow~\cite{ardizzone2019guided}, Auto-Regressive~\cite{Guadarrama17,royer2017probabilistic,baig2017multiple,kumar2021colorization,ji2022colorformer,weng2022ct,kang2023ddcolor}, GAN~\cite{cao17,yoo2019coloring,wu21,alghofaili2021exploring,zhao2021scgan,jin2021focusing,kim2022bigcolor,lee2022bridging,wang2022palgan}, and Diffusion Models~\cite{saharia2022palette,liu2023improved,zabari23}.

\textit{At present, Auto-Regressive (including Transformer) and GAN models are the mainstream generation architectures for designing diverse coloring methods. Since the generative algorithm is to model training data distribution, it has the inherent property of generating diverse samples. As the performance of the diffusion model shines in other vision tasks, its modeling performance for data diversity is superior to previous generative algorithms. Therefore, researchers tend to design multi-modal interaction-based coloring methods based on diffusion models.}

\begin{figure}[ht]
    \centering
    \includegraphics[width=1.0\linewidth]{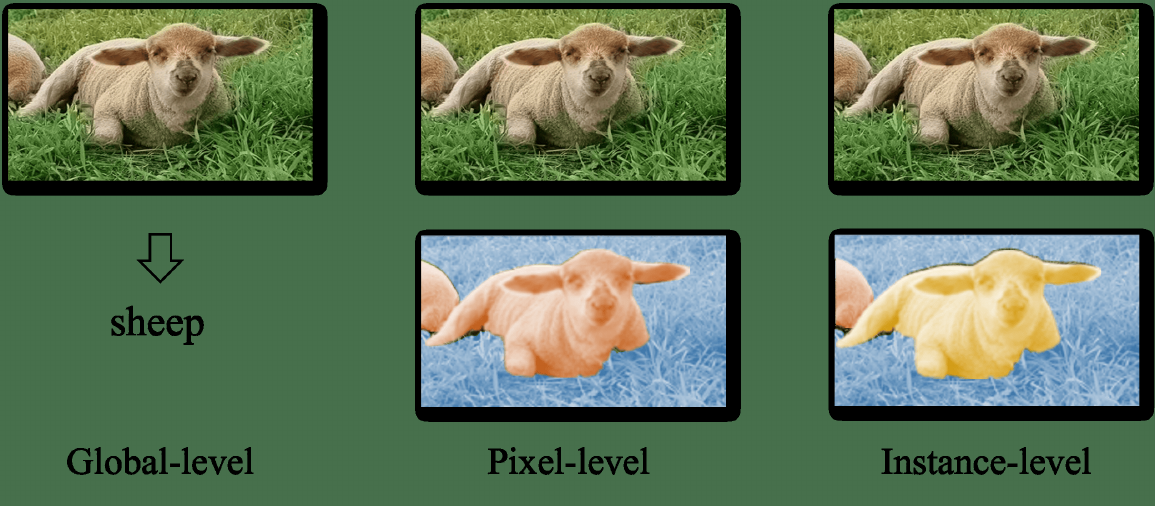}
    \vspace{-10px}
    \caption{Illustration of three kinds of automatic semantic colorization methods. Global level, pixel level, and instance level, separately correspond to three vision tasks: image classification, semantic segmentation, and instance segmentation.}
    \label{fig:semantic}
\end{figure}

{\flushleft \bf Semantic Colorization.}
Semantic information is essential for performing high-quality colorization. We divide the existing semantic automatic colorization methods into three categories, including global level, pixel level, and instance level, which correspond to three vision tasks: image classification, semantic segmentation, and instance segmentation, as is shown in Fig.~\ref{fig:semantic}.

{\flushleft 1. Global Level.} \\
Iizuka~et al.~\cite{iizuka2016let} was the first to propose using semantic information for a colorization process; their approach involves a fully automated, data-driven method to classify and colorize images utilizing a labeled dataset. It integrates local image features, calculated from local patches, with visual context derived from semantic class labels. This enables the model to determine and understand the correlation between semantic labels and their respective colors. 
Özbulak~\cite{ozbulak2019image} proposed a colorization approach, utilizing the image recognition and understanding capabilities of Capsule Networks (CapsNets)~\cite{sabour17}. 
Vitoria~et al.~\cite{vitoria2020chromagan} introduced an approach combining adversarial learning with semantic information for colorization. They leveraged the power of GAN to generate realistic and visually pleasing colored images. Their method incorporated semantic class distribution learning, guiding the colorization process based on the semantic context of the image.
The method proposed by Jin~et al.~\cite{jin2021broad} introduced a new automatic colorization approach based on a broad learning system~\cite{chen18}, which accurately determined the category and color of pixel blocks through training. 
Kang~et al.~\cite{kang2023ddcolor} proposed a semantically reasonable and visually vivid colorization approach. They utilized a pixel decoder and a query-based color decoder in their approach. The former was responsible for reconstructing the spatial information of the image, and the latter used rich visual cues to enhance color queries without requiring manually determined color priors.

{\flushleft 2. Pixel Level} \\
The image-level classification task typically prioritizes transform invariance, meaning the classification should remain the same even if the image is transformed. 
However, satisfactory colorization requires the use of transform-variant features. In this regard, semantic segmentation is viewed as a more appropriate sub-task for colorization, which similarly relies on pixel-level and transform-variant representations.
Following this approach, Zhao~et al.~\cite{zhao2018pixel} proposed a method incorporating detailed object semantics at the pixel level to direct the process of image colorization.
They trained a network with two loss functions, one for semantic segmentation and the other for colorization. By leveraging the pixel-level object semantics, their approach enhanced the precision and uniformity of the colorization process.
Building upon this work, Zhao~et al.~\cite{zhao2020pixelated} further investigated how to employ pixel-related semantics to produce appropriate colorization. In addition to segmenting objects with common illumination conditions, Duan~et al.~\cite{duan2024} proposed a method for colorizing shadowed images, which can be regarded as a specialized form of segmentation that distinguishes between shadowed and non-shadowed regions.

{\flushleft 3. Instance Level} \\
The various current approaches for learning semantics in colorization, whether at the image or at the pixel level, have limited ability to capture the variations in object appearance adequately. Although the above-reviewed models have demonstrated impressive results for a wide range of images, they struggled when faced with images containing multiple objects on cluttered backgrounds. 
To address this limitation, Su~et al.~\cite{su2020instance} proposed the first instance-aware colorization architecture incorporating an off-the-shelf object detector to capture segmented object visuals and also utilized an instance-based colorization network to extract features at the object level.
By leveraging this approach, the model could effectively handle complex scenes with multiple instances. 
Previous colorization methods have primarily focused on image-level or entity-level features, not adequately capturing how the object instances in an image interact with the overall context.
To address this limitation, Pucci~et al.~\cite{pucci2021collaborative} introduced UCapsNet with capsules~\cite{sabour17}, incorporating features at the image level produced by convolutions and features at the entity level.
Xia~et al.~\cite{xia2022disentangled} introduced the concept of disentangled colorization and proposed the identification of multiple color anchors that can effectively represent the diverse color distribution of an image. By specifying these anchor colors, their algorithm can predict the image colors, leveraging the global color affinity to ensure consistency in the overall structure. 
Chang~et al.~\cite{chang2023coins} built upon the concept of instance-awareness in colorization and introduced text guidance to establish a correspondence between instance regions and color descriptions. This allowed their model to assign colors to instances flexibly, even when such correspondences were not encountered during training. 

\textit{Colorization tasks benefit significantly from high-level image understanding, which has evolved towards finer granularity. Early methods treat the image as a whole, using image classification for global content understanding. This is referred to as the global level of understanding. Subsequently, methods evolved to the pixel level, where each pixel is understood through semantic segmentation. More recently, instance-level understanding has emerged, allowing for different color assignments to different instances.}

\subsubsection{Computer Vision Applications}
Automatic colorization can also support other applications in the computer vision field. Our survey mainly discusses two common applications: 1. computational imaging and 2. image restoration and enhancement.
{\flushleft \bf Computational Imaging.}
To improve the imaging quality of monochrome-color dual-lens systems,
Dong~et al.~\cite{dong2019learning} tackled the challenging reference-based colorization problem by expanding the number of pixels in the reference image and determining the color of each pixel in the input image as the ranked average of colors.
They used ResNet to capture high-level features from the input and reference images, refining the intermediate features and establishing feature relations using convolution and Sigmoid layers. 
As an extension of this work, Dong~et al.~\cite{dong2020colorization} introduced two parallel modules for colorization and colorization quality estimation, leveraging a colorization CNN. This work further advanced the techniques for improving the colorization process and had great practical value for improving the imaging quality of mobile devices in various applications.

{\flushleft \bf Image Restoration and Enhancement.}
In the CV field, some researchers proposed a general and multi-purpose vision model for several low-level image processing tasks and often evaluated the model performance in many applications, including colorization.
The seminal work by Ulyanov~et al.~\cite{ulyanov2018deep} demonstrated that a randomly initialized CNN implicitly captured texture-level image priors when trained iteratively on a single image, indicating that the CNN can be fine-tuned for image restoration tasks, such as reconstructing a corrupted image. 
Building on this, Pan~et al.~\cite{pan2021exploiting} expanded upon existing methods by utilizing image priors learned from a GAN trained on extensive natural images. This approach enabled flexible restoration and manipulation, unlike fixed generator assumptions in previous GAN inversion techniques. Their method exhibited strong generalization capabilities across various image restoration and manipulation tasks despite not being tailored to each specific task. It restored missing information while preserving semantic details when reconstructing corrupted images.
This demonstrated the potential for leveraging deep learning models to capture and apply rich image priors to diverse image restoration tasks.
Zhang~et al.~\cite{zhang2022scsnet} introduced an efficient SCSNet paradigm that used low-resolution gray-scale images as input and produced high-resolution colorful images using a unified model. Their model deployed colorization and super-resolution in two consecutive stages, the first stage incorporating a Pyramid Valve Cross Attention (PVCAttn) module to combine information regarding the source and reference images effectively. A Continuous Pixel Mapping (CPM) module was developed in the second stage to efficiently generate target images at any magnification, using discrete pixel correlation in a continuous space.
El Helou and Süsstrunk~\cite{el2022bigprior} viewed colorization as a generalization of classic restoration algorithms, such as image colorization, inpainting, and denoising. They decoupled the image restoration task into prior and data fidelity using an inversion GAN model, where a prior represented added information, such as the missing two-channel information for colorization.
Wang~et al.~\cite{wang2022zero} introduced a zero-shot framework designed for arbitrary linear image restoration problems, offering a versatile solution for diverse image restoration endeavors, including image super-resolution, colorization, inpainting, compressed sensing, and deblurring. 
In a related study, Chan~et al.~\cite{chan2023glean} investigated the latent bank of GAN to uncover existing natural image priors. They proposed a novel approach that utilized pre-trained GAN models for various image restoration tasks, including super-resolution, colorization, and hybrid restoration.

\subsection{Video Colorization}
The practice of hand-coloring photographs dates back to the inception of photography. Similarly, video colorization, which involves adding color to image sequences, has been a research subject from the outset of this field.
Yatziv and Sapiro~\cite{yatziv2006fast} proposed an approach that utilized an intrinsic, gradient-weighted distance measure to spread user-provided scribble colors across entire images or image sequences. Similarly, Jacob and Gupta~\cite{jacob2009colorization} also developed a method that relied heavily on user inputs regarding segmentation and colorization. Their method enhanced the process by using a keyframe to transfer color to other frames, employing a motion estimation algorithm. Sheng~et al.~\cite{sheng2013video} introduced a different approach that established pixel similarity in the video's gray-scale channel, allowing for parallel color optimization among pixels across video frames.

In the present deep learning era, we can broadly categorize existing video colorization methods into four types. The first type involves directly applying colorization over a single image and then post-processing the colorized image to achieve temporal consistency, as described previously in~\cite{bonneel2015blind}. This method ensures that the colorization remains consistent over time, providing a more realistic output. Another study~\cite{lei2020blind} presented a specific case demonstrating that training a model with Deep Video Prior can directly attain temporal consistency.

The second type of method performs colorization on individual frames and encodes temporal consistency using motion estimation~\cite{xia2016robust,liu2024temporally} or inter-frame similarity~\cite{liu2023video}. This approach ensures that the colorization remains consistent with the inter-frame correspondence throughout the video sequence. However, errors can accumulate due to inaccurate correspondence estimation.

The third type of method follows the tradition of example-based colorization, a colored frame being used as a reference, and the colorization of subsequent frames is being achieved by evaluating content similarity~\cite{ben2015approximate,lai2018learning,liu2018switchable,vondrick2018tracking,lei2019fully,casey2021animation,shi2023reference,zhao2023svcnet}. This method ensures a consistent color scheme throughout the video based on the reference frame. Additionally, other methods within this type employed information propagation techniques to colorize the video frames~\cite{jampani2017video,meyer2018deep,zhang2019deep,zhang2021line,yang2022bistnet}. This approach allows for transferring color information from one frame to another, ensuring a coherent colorization across the video.

The fourth and the last type of methods consider spatiotemporal features directly~\cite{paul2016spatiotemporal,chen2024exemplar}, which eliminates the need for motion estimation or sequential information transfer from frame to frame, making it a more efficient method for video colorization.

\textit{Video colorization presents a unique challenge compared to single image colorization: striking a balance between maintaining spatiotemporal color coherence and managing processing costs.}

\begin{table*}[th!]
\centering
\caption{An Overview of Major Datasets Used for Training Networks in Colorization Tasks}
\begin{tabular}{ccccc}
\rowcolor{gray!10}
\toprule
 Category & Dataset & Year & Type & Paper Used \\  
\hline
\rowcolor{red!10}
                                  & CIFAR~\cite{cifar09} & 2009 & Image classification &~\cite{royer2017probabilistic,baig2017multiple} \\ \rowcolor{red!10}              
                                  & ImageNet~\cite{imagenet09} & 2009 & Visual recognition & \begin{tabular}[c]{@{}l@{}}~\cite{he18,xu20,lu20,yin21,bai2022semantic,wang23,richard17,xiao22,yun23,Guadarrama17}\\ ~\cite{royer2017probabilistic,baig2017multiple,Deshpande17,messaoud18,wu21,lei2019fully,zhao2023svcnet,zhang2019deep,zhang2016colorful}\end{tabular}\\ \rowcolor{red!10}
                                  & PASCAL VOC~\cite{pascal2010} & 2010 & Object recognition &~\cite{li21,zhao2018pixel,zhao2020pixelated}\\ \rowcolor{red!10}
                                  & COCO~\cite{coco14} & 2014 & Object understanding &~\cite{xu20,Carrillo_2022_ACCV,manjunatha2018learning} \\ \rowcolor{red!10}
                                  & Places 205~\cite{places205} & 2014 & Scene recognition &~\cite{li21,iizuka2016let,jin2021broad}\\ \rowcolor{red!10}
                                  & LSUN~\cite{lsun15} & 2015 & Scene understanding &~\cite{Deshpande17,cao17,messaoud18} \\ \rowcolor{red!10}
\multirow{-7}{*}{gray-scale Image} & ADE20K~\cite{ade20k17} & 2017 & Semantic segmentation &~\cite{xiao2020example} \\ 
\hline
\rowcolor{yellow!10}
                               & MIT-Adobe 5K~\cite{by2011} & 2011  & Image enhancement &~\cite{chang2015palette} \\ \rowcolor{yellow!10}
\multirow{-2}{*}{Color Platte} & Color Theme~\cite{wang2012affective} & 2012 & Image colorization &~\cite{wang2012affective} \\ 
\hline
\rowcolor{blue!10}
                               & CoSaL~\cite{chen2018language} & 2018  & Image colorization &~\cite{chen2018language} \\ \rowcolor{blue!10}
                               & Palette-and-Text~\cite{bahng2018coloring} & 2018 & Image colorization &~\cite{bahng2018coloring} \\ \rowcolor{blue!10}
                               & SketchyScene~\cite{zou18} & 2018 & Sketch colorization &~\cite{zou2019language} \\ \rowcolor{blue!10}
                               & Extended COCO-stuff~\cite{weng2022code} & 2022 & Image colorization &~\cite{weng2022code,chang2022coder,chang2023coins,chang2023cad} \\ \rowcolor{blue!10}
\multirow{-5}{*}{Text Guidance} & Multi-instance~\cite{chang2023coins} & 2023 & Image colorization  &~\cite{chang2023coins,chang2023cad} \\
\hline
\rowcolor{green!10}
                        & DAVIS~\cite{davis16} & 2016 & Video object segmentation &~\cite{lai2018learning,lei2019fully,liu2024temporally,zhao2023svcnet,meyer2018deep,yang2022bistnet}\\ \rowcolor{green!10}
                        & Kinetics~\cite{kinetics17} & 2017 & Human action recognition &~\cite{vondrick2018tracking} \\ \rowcolor{green!10}
\multirow{-3}{*}{Video} & Videvo~\cite{videvo} & 2018 & Free stock video footage &~\cite{lai2018learning,liu2024temporally,zhao2023svcnet,zhang2019deep,yang2022bistnet} \\

\hline
\rowcolor{brown!10}
                               & Manga109~\cite{manga109-17} & 2017 & Media processing of manga &~\cite{fu17} \\ \rowcolor{brown!10}
                               & Danbooru2018~\cite{danboorucharacter} & 2018 & Anime character recognition &~\cite{akita2020colorization,ci18,zhang18,yuan2021line,zhang2021user,dou21,carrillo2023diffusart,animediffusion24} \\ \rowcolor{brown!10}
                               & Anime Sketch Colorization Pair~\cite{ascp} & 2019 & Sketch colorization &~\cite{liu2022reference,cao23,wu23} \\ \rowcolor{brown!10}
\multirow{-4}{*}{Line Drawing} & Tag2Pix~\cite{kim2019tag2pix} & 2019 & Sketch colorization &~\cite{kim2019tag2pix,cao2021line,cho23}\\                                                                           
\bottomrule
\end{tabular}
\label{tbl:dataset}

\end{table*}

\section{Datasets}\label{sec:dataset}
There are two situations in which researchers use datasets for colorization tasks: one for testing the performance of traditional methods where researchers search the website for images and the other for training networks using large-scale image datasets when designing deep learning-based models. This survey reviews the datasets used for \textbf{training} learning-based colorization models, as captured in Table~\ref{tbl:dataset}. 
Datasets have been sorted chronologically according to the publication date and classified according to the types of input to be colorized. ImageNet~\cite{imagenet09} is the most used large-scale dataset in natural gray-scale image colorization, and researchers use this dataset to train models to acquire the color prior. Most of these gray-scale image datasets are mainly used for other vision tasks, yet they are also suitable for colorization tasks due to their detailed annotations and scene categories. Researchers can obtain many data pairs by converting the original color images into gray-scale images. For video colorization, researchers mainly use DAVIS~\cite{davis16} and Videvo~\cite{videvo} datasets to train their models, especially for learning spatiotemporal consistency in videos. Unlike natural images that can be captured easily on a large scale, line drawing data are challenging to collect. There are two solutions to acquire such data: the first one is to invite professional painters to draw manually, which is relatively expensive and time-consuming, and the other is to extract the lines from the colored animation data using SketchKeras~\cite{sketchKeras} or Anime2Sketch~\cite{Anime2Sketch}, although the latter has very different line types than actual hand-painted line drawings. In addition, the Danbooru dataset~\cite{danboorucharacter} is a valuable resource for anyone interested in anime-style artwork. The most notable feature is its extensive tagging system. Images are annotated with a wide range of tags describing various aspects of the content, such as characters, themes, styles, and specific elements. This rich tagging system makes the dataset particularly valuable for training machine learning models in tasks like image classification, object detection, style transfer, and colorization.

\begin{figure*}[ht]
    \centering
    \includegraphics[width=1.0\linewidth]{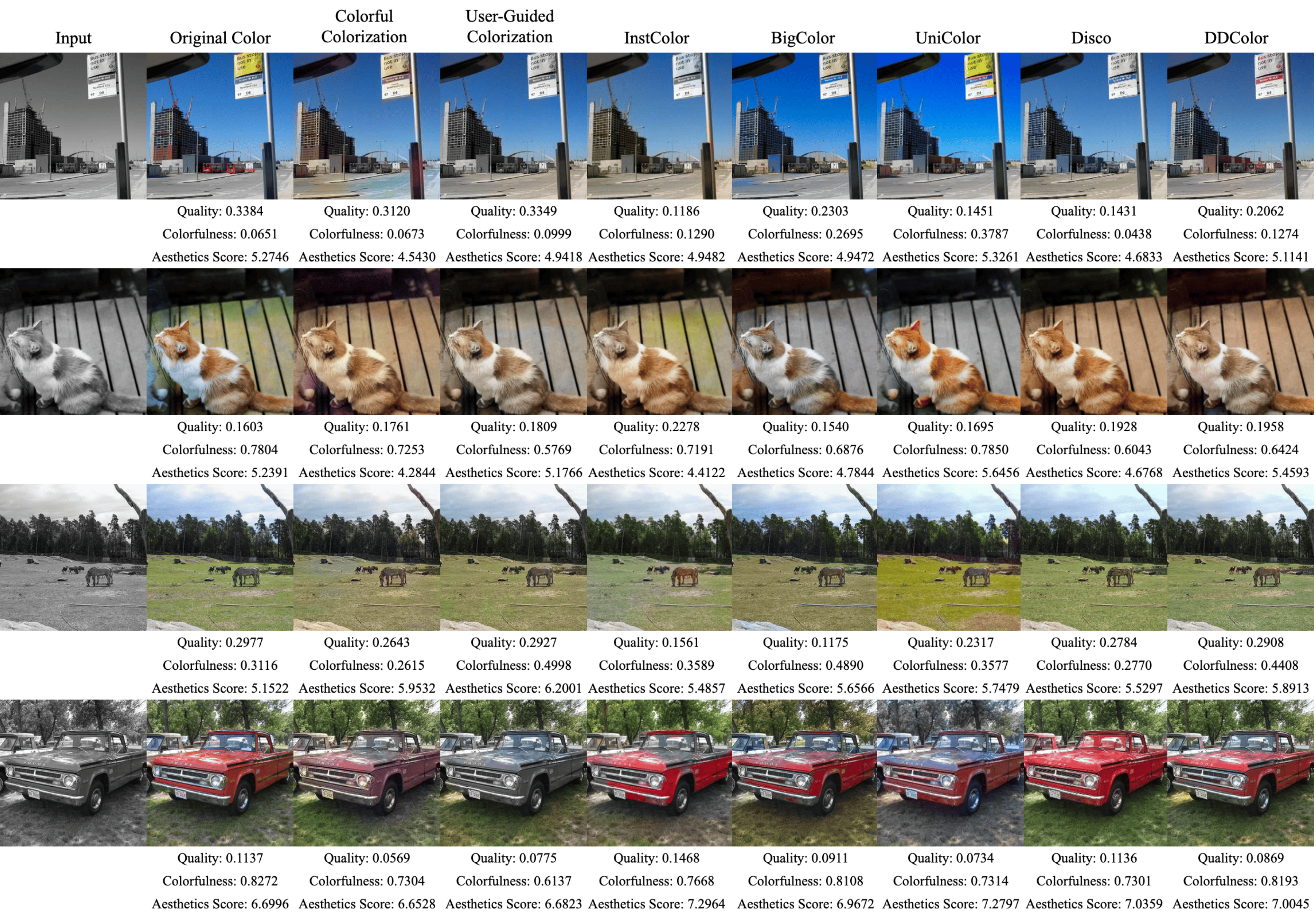}
    \vspace{-10px}
    \caption{Colorization aesthetic assessment of seven automatic colorization models, including Colorful Colorization~\cite{zhang2016colorful}, User-Guided Colorization~\cite{richard17}, InstColor~\cite{su2020instance}, BigColor~\cite{kim2022bigcolor}, UniColor~\cite{huang2022unicolor}, Disco~\cite{xia2022disentangled}, and DDColor~\cite{kang2023ddcolor}. Under each example of the original color image and the colored image, we show three scores computed by CLIP-IQA~\cite{wang2023exploring} and LAION-Aesthetics Predictor V2~\cite{laion}. The first two scores evaluate image quality from the image's overall texture and color dimensions, while the last score evaluates image quality from the aesthetic aspect. It can be seen that the colorization results of some samples exceed the original color images in terms of aesthetic scores.}
    \label{fig:comp}
\end{figure*}

\begin{figure*}[h]
    \centering
    \includegraphics[width=1.0\linewidth]{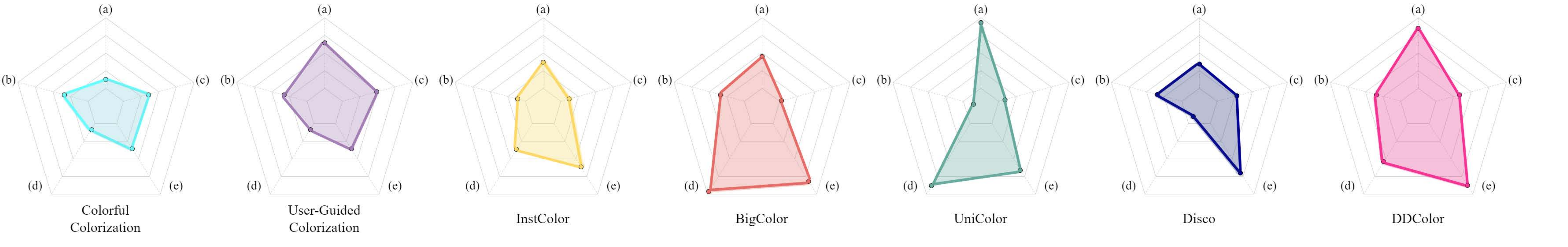}
    \vspace{-10px}
    \caption{Visualization of model performance of seven automatic colorization methods, including Colorful Colorization~\cite{zhang2016colorful}, User-Guided Colorization~\cite{richard17}, InstColor~\cite{su2020instance}, BigColor~\cite{kim2022bigcolor}, UniColor~\cite{huang2022unicolor}, Disco~\cite{xia2022disentangled}, and DDColor~\cite{kang2023ddcolor}. (a) Aesthetic Score, (b) Inference Time, (c) Quality, (d) Colorfulness, and (e) FID. All values are the normalized average scores.}
    \label{fig:vis}
\end{figure*}

\section{Colorization Aesthetic Assessment }\label{sec:metric}
Assessing the performance of various colorization methods is inherently linked to human visual aesthetic perception.
A hybrid evaluation methodology incorporating subjective and objective assessments is standard in the published literature.
Subjective evaluation involves qualitative analysis provided by the model designers and user studies based on participants' visual perceptions. 
Commonly used objective evaluation metrics for assessing colorization performance include PSNR, SSIM, LPIPS, and FID. Huang~et al.~\cite{HUANG2022105006} provided a detailed analysis of the existing evaluation metrics used in colorization.
However, these first three reconstruction-based evaluation metrics were initially designed for image quality assessment~\cite{wang2004} rather than for colorization quality assessment. They evaluate the colorization performance by calculating the difference between the colored image and the original color image of the input gray-scale image, i.e., the generated image should be as close to the ground-truth image as possible.
Given that the colored rendition of a gray-scale image lacks uniqueness, achieving an exact replication of the original color image is deemed unnecessary.
There should not be a ground-truth output for any given input gray-scale image.
Most learning-based methods train the colorization networks with paired training data. As a result, the mainstream evaluation method compares the colorization results with ground-truth color images in the test dataset.
In addition, FID is a metric used to evaluate the performance of generative models. Although some generative model-based colorization methods may exhibit superior generation ability, this does not necessarily mean that their colorization performance aligns with the generation abilities. 

In this paper, in addition to reviewing the current state-of-the-science, we explore using a neural network model to simulate human visual perception to evaluate the performance of \textbf{automatic colorization models.}
Inspired by the work of image aesthetic assessment~\cite{deng2017}, we propose a new concept of \textit{colorization aesthetic assessment.} 
Human aesthetic evaluation of colors involves psychology and visual cognition elements, which hold significant practical value in computational imaging. For example, when developing imaging algorithms for mobile phone photography, different cell phone manufacturers apply distinctly different color adjustments, resulting in different color effects in photos of the same scene as other mobile phone brands. 
The key problem of aesthetic evaluation is quantifying subjective judgment, so we use two metrics based on large-scale cross-modal data training to conduct our coloring aesthetic evaluation test. 
In this survey, we mainly focus on testing the unconditional natural gray-scale colorization performance since the automatic coloring method can reflect the model's ability for semantic understanding and color representation.
As a demonstration, the two metrics are used to evaluate the following seven representative colorization models: Colorful Colorization~\cite{zhang2016colorful}, User-Guided Colorization~\cite{richard17}, InstColor~\cite{su2020instance}, BigColor~\cite{kim2022bigcolor}, UniColor~\cite{huang2022unicolor}, Disco~\cite{xia2022disentangled}, and DDColor~\cite{kang2023ddcolor}.
All models use automatic colorization methods or are set in automatic mode, which means no user-guided color information is included.

\begin{table*}[h]
	\centering
	\caption{Quantitative Comparison of Seven Automatic Colorization Methods Using Existing Evaluation Metrics}
	\begin{tabular}{ccccccccc}
		\toprule
        \multirow{2}{*}{Method} & \multicolumn{4}{c}{COCO test dataset (5k)} & \multicolumn{4}{c}{ImageNet test dataset (10k)} \\
                                \cmidrule(r){2-5} \cmidrule(r){6-9} 
                                & PSNR$\uparrow$ & SSIM$\uparrow$ & LPIPS$\downarrow$ & FID$\downarrow$ & PSNR$\uparrow$ & SSIM$\uparrow$ & LPIPS$\downarrow$ & FID$\downarrow$ \\
		\midrule 
        Colorful Colorization~\cite{zhang2016colorful} & 22.96 & 0.998 & 0.213 & 17.90 & 22.84 & 0.998 & 0.219 & 9.72 \\
        User-Guided Colorization~\cite{richard17} & 24.75 & 0.998 & 0.171 & 16.82 & 24.67 & 0.998 & 0.180 & 8.55 \\
        InstColor~\cite{su2020instance} & 22.35 & 0.838 & 0.238 & 12.72 & 22.03 & 0.909 & 0.217 & 7.35 \\
        BigColor~\cite{kim2022bigcolor} & 21.40 & 0.887 & 0.214 & 9.05 & 21.57 & 0.886 & 0.212 & 3.67 \\
        UniColor~\cite{huang2022unicolor} & 23.09 & 0.912 & 0.202 & 11.16 & 21.73 & 0.909 & 0.236 & 6.93 \\
        Disco~\cite{xia2022disentangled} & 20.46 & 0.851 & 0.236 & 10.59 & 20.72 & 0.862 & 0.229 & 5.57 \\
        DDColor~\cite{kang2023ddcolor} & 23.41 & 0.997 & 0.184 & 7.88 & 23.18 & 0.997 & 0.192 & 3.93 \\
		\bottomrule  
	\end{tabular}
    \label{tbl:comp}
\end{table*}

\begin{table}[h]
	\centering
	\caption{Inference Time Comparison of Seven Automatic Colorization Methods}
	\begin{tabular}{cc}
		\toprule
            Method & Time cost (seconds) \\
		\midrule 
        Colorful Colorization~\cite{zhang2016colorful} & 0.0749 \\
        User-Guided Colorization~\cite{richard17} & 0.0853 \\
        InstColor~\cite{su2020instance} & 0.8144   \\
        BigColor~\cite{kim2022bigcolor} & 0.0875  \\
        UniColor~\cite{huang2022unicolor} & 1.7162  \\
        Disco~\cite{xia2022disentangled} & 0.0945\\
        DDColor~\cite{kang2023ddcolor} & 0.0779\\
		\bottomrule  
	\end{tabular}
    \label{tbl:infertime}
\end{table}

These two metrics are briefly described below.
The first metric, CLIP-IQA~\cite{wang2023exploring}, represents an image quality assessment metric, leveraging the visual language prior embedded in CLIP~\cite{radford2021learning} model. This metric enables the evaluation of both the perceptual quality and abstract perception of images without the need for explicit task-specific training.
Rather than computing the cosine similarity between two entire feature vectors of image and text, the cosine similarity between the image feature vector $x$ and every antonym prompt pair $(t_1, t_2)$ is instead first computed to reduce the ambiguity of text prompts as follows:
\begin{equation}
\label{equ:1}
     s_i = \frac{x \bigodot t_i}{\lVert x \rVert \cdot \lVert t_i \rVert}, i \in \{1, 2\},
\end{equation}
Softmax is then used to compute the final score $\overline{s} \in [0, 1]$ as follows:
\begin{equation}
\label{equ:2}
     \overline{s} = \frac{e^{s_1}}{e^{s_1} + e^{s_2}}.
\end{equation}
The CLIP-IQA metric can assess overall and fine-grained image quality and contains many built-in evaluation prompts. We mainly use `quality' and `colorfulness' to evaluate colorized images, including quality and color perceptions. Here, `quality' refers to the overall picture quality of the image, and 'colorfulness' measures how rich and saturated the colors appear.
The higher the score, the higher the image quality, contributing to the image's overall visual impact and aesthetic appeal.

The second metric, LAION-Aesthetics Predictor V2~\cite{laion}, is an image aesthetic assessment metric that takes CLIP image embeddings produced using the clip-vit-large-patch14 model as input and then outputs the aesthetics score ranging from 1 to 10, by concatenating a simple linear model. It is trained on the following three datasets: SAC, LAION-Logos, and AVA. The higher the score, the higher the aesthetic quality.

We perform non-reference image quality and aesthetic assessments to seven representative colorization methods using COCO test dataset~\cite{coco14} with central cropped testing images of resolution $256\times 256$. As illustrated in Fig.~\ref{fig:comp}, these seven colorization methods accurately represent the global semantic analysis of gray-scale images. The color prior learned through the training on a large number of natural image datasets can identify the sky as blue and the lawn as green. However, the automatic colorization method is unsuitable for controlling the details, such as the bus in the first row of Fig.~\ref{fig:comp}. An interactive approach will work better for this fine-grained coloring task. In addition, the examples in the second and fourth rows of Fig.~\ref{fig:comp} also reveal pattern collapse problems when training with paired data. The seven methods tend to consider the patterned fur coat of the cat to be orange and the car's paint to be red.

Regarding aesthetic evaluation, UniColor~\cite{huang2022unicolor} produces high saturation coloring results and achieves the highest score in color and aesthetics. In particular, the cat example displays a prominent cat body coloring effect, which is more beautiful than the original image. BigColor~\cite{kim2022bigcolor} can also produce a high saturation coloring effect.
InstColor~\cite{su2020instance}, Disco~\cite{xia2022disentangled}, and DDColor~\cite{kang2023ddcolor} produce a similar color effect, which is relatively soft. In terms of score, DDColor performs slightly better among the three. Colorful Colorization~\cite{zhang2016colorful} and User-Guided Colorization~\cite{richard17} are two early learning-based methods. Their results have the problems of color bleeding (the third column of the first row) or incomplete coloring (the fourth column of the last row) on individual examples. According to the experimental data, the quality score is not positively correlated with the aesthetic score. Some samples have high aesthetic scores but low-quality scores.
This is because coloring is a process of image recovery and information enhancement for gray-scale images, and different methods have different processing mechanisms for gray-scale images. Hence, the information gain and image loss are different. 
As is shown in Fig.~\ref{fig:vis}, We used a radar map to visualize indicators in five dimensions, including (a) Aesthetic Score, (b) Inference Time, (c) Quality, (d) Colorfulness, and (e) FID. All the values are normalized, and the larger the area, the better the comprehensive performance of the method.

As illustrated in Table~\ref{tbl:comp}, we also evaluate automatic colorization methods using traditional assessment metrics on the COCO test and ImageNet test dataset. All images are centrally cropped to the resolution of $256\times 256$ instead of directly resizing to maintain the object structure without geometry distortion.
The reconstruction-based evaluation metrics and generation ability index can not intuitively reflect the performance of the colorization models for color information representation. The FID scores illustrate that the early methods~\cite{zhang2016colorful,richard17} without using generative models show worse modeling ability for the diversity of data samples than other methods in both test datasets.
In Table~\ref{tbl:infertime}, we compare the inference time of seven automatic models. All models were tested on a single RTX 4090 GPU, and the values shown in the table are the average time taken to colorize each image. After a comprehensive evaluation, DDColor~\cite{kang2023ddcolor} is the current most SOTA automatic gray-scale image colorization method.
Using the theoretical aspects of vision to evaluate the content produced by graphics is very much in line with the current research concept of vision for graphics, such as assessing the generated contents of AIGC models.
We hope our initial attempts to perform a colorization aesthetic assessment will inspire the research community to pursue such an aesthetic approach and understanding in future research. This will significantly simplify and reduce the labor-intensive and burdensome nature of colorization in the future. 

\section{Discussion and Future Work}\label{sec:discussion}
After over two decades of development of colorization technology, from the initial user-guided gray-scale image rendering technology, it gradually evolved to more advanced learning-based generative multi-modal colorization. The coloring targets have varied, covering images and videos of gray-scale and line art. In this section, we give the directions that are still valuable in the study of colorization and need to be explored further, hoping to inspire subsequent researchers.
{\flushleft \bf Integrate colorization with AIGC technology.}
With the explosion of AIGC technology, the task of colorization revolutionized. For example, in reference-based line drawing colorization, ControlNet~\cite{zhang2023adding}-based works provide novel solutions that formulate the problem as a line drawing guided reference image editing instead of injecting color from reference image to target line drawing. Directly editing the colored artwork to create identify-consistent results will significantly improve the efficiency of animation creation. Integrating the colorization method with AIGC technology will produce more diverse results for users to refer to and choose the appropriate scheme. 

{\flushleft \bf The trend of multi-modal interaction method.}
At the beginning of the development of coloring technology, the methods proposed by researchers were based on user guidance, and various user input forms were derived. Then, using a single form is always insufficient, so multi-modal fusion is worthy of in-depth study. These pioneering multi-modal methods~\cite{huang2022unicolor,liang2024control,yan2024colorizediffusion} verify the effectiveness of the multi-modal coloring technique. Human guidance and interaction are still essential in the coloring process, and the neural network model only helps us fill in some colors prior. More details still need to be manually fine-tuned to generate final plausible results.

{\flushleft \bf Generalization ability of reference-based method.}
The reference-based method, especially for line drawing colorization, has been developed so far, and a variety of works~\cite{lee20,li2022eliminating,cao23,animediffusion24,lin24} have been proposed. Still, cross-domain feature matching is the core problem that always needs to be solved. Since the lack of a paired training dataset, Lee~et al.~\cite{lee20} proposed the self-supervised augmented training strategy as a compromise that formulated the problem of line drawing guided image restoration for learning the cross-domain feature matching. However, this method's generalization performance is unsuitable for conducting accurate colorization. Different from real-world images, the feature correspondence of cartoons needs to be designed separately and trained with customized datasets.

{\flushleft \bf Image and Video Editing.}
Colorization techniques can be used for image and video editing, such as recolorization or color transfer~\cite{zhao21,ke23}.
In addition to the editing of color information in photographs and videos, the academic community is increasingly interested in more fancy research, such as non-photorealistic artistic rendering of images~\cite{ebsynth} or videos~\cite{yang23,qi23}. In particular, with the help of text control, a variety of artistic editing can be carried out on the screen content, which has strong playability and commercial value. As an extension of traditional coloring techniques, how to execute these `magic applications' with low power consumption on the end side so that users can realize these operations more conveniently on the mobile phone is an essential direction of this kind of research in the future.

{\flushleft \bf Dedicated new assessment method for colorization.}
To evaluate the performance of colorization methods, researchers utilized different assessment methods, including subject and object evaluation metrics. Huang~et al.~\cite{HUANG2022105006} detailed the traditional evaluation metrics used in the previous research. In this paper, we initially attempt to introduce aesthetic assessment into colorization, and it is still an open, challenging question to design an evaluation tailored for colorization tasks. As the assessment of coloring results is highly subjective, and aesthetics is one of the aspects, a more comprehensive coloring evaluation is a precious research direction.

\section{Conclusions}
\label{sec:summary}
This survey provides a comprehensive overview of the research and state-of-the-science in colorization technologies, which is considered an application research in computer graphics. Based on this view, we offer a new taxonomy of colorization technology and detail the subcategories and contents. Based on our systematic investigation and analysis, we conclude that colorization originates from graphics, prospers in vision, and forms a fusion, i.e., \textit{vision for graphics}, with the rapid development of generative models. In the process, we also extend current colorization evaluation metrics and propose the concept of colorization aesthetic assessment for evaluating seven automatic colorization methods.
We explore the challenges and potential future directions of colorization tasks, and we hope this survey serves as a valuable resource for future researchers in the field of colorization research.

\ifCLASSOPTIONcaptionsoff
  \newpage
\fi

\bibliographystyle{IEEEtran}
\bibliography{reference}

\begin{thebibliography}{100}
\providecommand{\url}[1]{#1}
\csname url@samestyle\endcsname
\providecommand{\newblock}{\relax}
\providecommand{\bibinfo}[2]{#2}
\providecommand{\BIBentrySTDinterwordspacing}{\spaceskip=0pt\relax}
\providecommand{\BIBentryALTinterwordstretchfactor}{4}
\providecommand{\BIBentryALTinterwordspacing}{\spaceskip=\fontdimen2\font plus
\BIBentryALTinterwordstretchfactor\fontdimen3\font minus \fontdimen4\font\relax}
\providecommand{\BIBforeignlanguage}[2]{{%
\expandafter\ifx\csname l@#1\endcsname\relax
\typeout{** WARNING: IEEEtran.bst: No hyphenation pattern has been}%
\typeout{** loaded for the language `#1'. Using the pattern for}%
\typeout{** the default language instead.}%
\else
\language=\csname l@#1\endcsname
\fi
#2}}
\providecommand{\BIBdecl}{\relax}
\BIBdecl

\bibitem{iizuka2016let}
S.~Iizuka, E.~Simo-Serra, and H.~Ishikawa, ``Let there be color!: joint end-to-end learning of global and local image priors for automatic image colorization with simultaneous classification,'' \emph{ACM Transactions on Graphics}, vol.~35, no.~4, pp. 110:1--110:11, 2016.

\bibitem{vitoria2020chromagan}
P.~Vitoria, L.~Raad, and C.~Ballester, ``{ChromaGAN}: Adversarial picture colorization with semantic class distribution,'' in \emph{IEEE Winter Conference on Applications of Computer Vision}, 2020, pp. 2434--2443.

\bibitem{jin2021broad}
Y.~Jin, B.~Sheng, P.~Li, and C.~L.~P. Chen, ``Broad colorization,'' \emph{IEEE Transactions on Neural Networks and Learning Systems}, vol.~32, no.~6, pp. 2330--2343, 2021.

\bibitem{xia2022disentangled}
M.~Xia, W.~Hu, T.-T. Wong, and J.~Wang, ``Disentangled image colorization via global anchors,'' \emph{ACM Transactions on Graphics}, vol.~41, no.~6, pp. 204:1--204:13, 2022.

\bibitem{kang2023ddcolor}
X.~Kang, T.~Yang, W.~Ouyang, P.~Ren, L.~Li, and X.~Xie, ``{DDColor}: Towards photo-realistic image colorization via dual decoders,'' in \emph{IEEE International Conference on Computer Vision}, 2023, pp. 328--338.

\bibitem{zhao2018pixel}
J.~Zhao, L.~Liu, C.~G.~M. Snoek, J.~Han, and L.~Shao, ``Pixel-level semantics guided image colorization,'' in \emph{British Machine Vision Conference}, 2018, pp. 1--12.

\bibitem{ozbulak2019image}
G.~Özbulak, ``Image colorization by capsule networks,'' in \emph{IEEE Conference on Computer Vision and Pattern Recognition Workshops}, 2019, pp. 2150--2158.

\bibitem{sabour17}
S.~Sabour, N.~Frosst, and G.~E. Hinton, ``Dynamic routing between capsules,'' in \emph{Advances in Neural Information Processing Systems}, vol.~30, 2017, pp. 3856--3866.

\bibitem{zhao2020pixelated}
J.~Zhao, J.~Han, L.~Shao, and C.~G.~M. Snoek, ``Pixelated semantic colorization,'' \emph{International Journal of Computer Vision}, vol. 128, pp. 818--834, 2020.

\bibitem{su2020instance}
J.-W. Su, H.-K. Chu, and J.-B. Huang, ``Instance-aware image colorization,'' in \emph{IEEE Conference on Computer Vision and Pattern Recognition}, 2020, pp. 7965--7974.

\bibitem{pucci2021collaborative}
R.~Pucci, C.~Micheloni, and N.~Martinel, ``Collaborative image and object level features for image colourisation,'' in \emph{IEEE Conference on Computer Vision and Pattern Recognition Workshops}, 2021, pp. 2160--2169.

\bibitem{chang2023coins}
Z.~Chang, S.~Weng, P.~Zhang, Y.~Li, S.~Li, and B.~Shi, ``{L-CoIns}: Language-based colorization with instance awareness,'' in \emph{IEEE Conference on Computer Vision and Pattern Recognition}, 2023, pp. 19\,221--19\,230.

\bibitem{chen2018language}
J.~Chen, Y.~Shen, J.~Gao, J.~Liu, and X.~Liu, ``Language-based image editing with recurrent attentive models,'' in \emph{IEEE Conference on Computer Vision and Pattern Recognition}, 2018, pp. 8721--8729.

\bibitem{lstm}
S.~Hochreiter and J.~Schmidhuber, ``Long short-term memory,'' \emph{Neural Computation}, vol.~9, no.~8, p. 1735–1780, 1997.

\bibitem{bahng2018coloring}
H.~Bahng, S.~Yoo, W.~Cho, D.~K. Park, Z.~Wu, X.~Ma, and J.~Choo, ``Coloring with words: Guiding image colorization through text-based palette generation,'' in \emph{European Conference on Computer Vision}, 2018, pp. 443--459.

\bibitem{manjunatha2018learning}
V.~Manjunatha, M.~Iyyer, J.~Boyd-Graber, and L.~Davis, ``Learning to color from language,'' in \emph{North American Chapter of the Association for Computational Linguistics}, 2018, pp. 764--769.

\bibitem{kim2019tag2pix}
H.~Kim, H.~Y. Jhoo, E.~Park, and S.~Yoo, ``{Tag2Pix}: Line art colorization using text tag with {SECat} and changing loss,'' in \emph{IEEE International Conference on Computer Vision}, 2019, pp. 9055--9064.

\bibitem{zou2019language}
C.~Zou, H.~Mo, C.~Gao, R.~Du, and H.~Fu, ``Language-based colorization of scene sketches,'' \emph{ACM Transactions on Graphics}, vol.~38, no.~6, pp. 233:1--233:16, 2019.

\bibitem{cao2021line}
R.~Cao, H.~Mo, and C.~Gao, ``Line art colorization based on explicit region segmentation,'' \emph{Computer Graphics Forum}, vol.~40, no.~7, pp. 1--10, 2021.

\bibitem{weng2022code}
S.~Weng, H.~Wu, Z.~Chang, J.~Tang, S.~Li, and B.~Shi, ``{L-CoDe}: Language-based colorization using color-object decoupled conditions,'' in \emph{AAAI Conference on Artificial Intelligence}, vol.~36, no.~3, 2022, pp. 2677--2684.

\bibitem{chang2022coder}
Z.~Chang, S.~Weng, Y.~Li, S.~Li, and B.~Shi, ``{L-CoDer}: Language-based colorization with color-object decoupling transformer,'' in \emph{European Conference on Computer Vision}, 2022, pp. 360--375.

\bibitem{chang2023cad}
{Z. Chang, S. Weng, P. Zhang, Y. Li, S. Li, and B. Shi}, ``{L-CAD}: Language-based colorization with any-level descriptions using diffusion priors,'' in \emph{Advances in Neural Information Processing Systems}, vol.~36, 2023, pp. 77\,174--77\,186.

\bibitem{rombach2022high}
R.~Rombach, A.~Blattmann, D.~Lorenz, P.~Esser, and B.~Ommer, ``High-resolution image synthesis with latent diffusion models,'' in \emph{IEEE Conference on Computer Vision and Pattern Recognition}, 2022, pp. 10\,674--10\,685.

\bibitem{zhang2022scsnet}
J.~Zhang, C.~Xu, J.~Li, Y.~Han, Y.~Wang, Y.~Tai, and Y.~Liu, ``{SCSNet:} an efficient paradigm for learning simultaneously image colorization and super-resolution,'' in \emph{AAAI Conference on Artificial Intelligence}, vol.~36, no.~3, 2022, pp. 3271--3279.

\bibitem{el2022bigprior}
M.~El~Helou and S.~Süsstrunk, ``{BIGPrior}: Towards decoupling learned prior hallucination and data fidelity in image restoration,'' \emph{IEEE Transactions on Image Processing}, vol.~31, pp. 1628--1640, 2022.

\bibitem{pan2021exploiting}
X.~Pan, X.~Zhan, B.~Dai, D.~Lin, C.~C. Loy, and P.~Luo, ``Exploiting deep generative prior for versatile image restoration and manipulation,'' \emph{IEEE Transactions on Pattern Analysis and Machine Intelligence}, vol.~44, no.~11, pp. 7474--7489, 2021.

\bibitem{wang2022zero}
Y.~Wang, J.~Yu, and J.~Zhang, ``Zero-shot image restoration using denoising diffusion null-space model,'' in \emph{International Conference on Learning Representations}, 2023, pp. 1--31.

\bibitem{chan2023glean}
K.~C. Chan, X.~Xu, X.~Wang, J.~Gu, and C.~C. Loy, ``{GLEAN}: Generative latent bank for image super-resolution and beyond,'' \emph{IEEE Transactions on Pattern Analysis and Machine Intelligence}, vol.~45, no.~3, pp. 3154--3168, 2023.

\bibitem{ulyanov2018deep}
D.~Ulyanov, A.~Vedaldi, and V.~Lempitsky, ``Deep image prior,'' in \emph{IEEE Conference on Computer Vision and Pattern Recognition}, 2018, pp. 9446--9454.

\bibitem{santhanam2017generalized}
V.~Santhanam, V.~I. Morariu, and L.~S. Davis, ``Generalized deep image to image regression,'' in \emph{IEEE Conference on Computer Vision and Pattern Recognition}, 2017, pp. 5395--5405.

\bibitem{isola2017image}
P.~Isola, J.-Y. Zhu, T.~Zhou, and A.~A. Efros, ``Image-to-image translation with conditional adversarial networks,'' in \emph{IEEE Conference on Computer Vision and Pattern Recognition}, 2017, pp. 5967--5976.

\bibitem{zhang2016colorful}
R.~Zhang, P.~Isola, and A.~A. Efros, ``Colorful image colorization,'' in \emph{European Conference on Computer Vision}, 2016, pp. 649--666.

\bibitem{larsson2016learning}
G.~Larsson, M.~Maire, and G.~Shakhnarovich, ``Learning representations for automatic colorization,'' in \emph{European Conference on Computer Vision}, 2016, pp. 577--593.

\bibitem{larsson2017colorization}
{G. Larsson, M. Maire, and G. Shakhnarovich}, ``Colorization as a proxy task for visual understanding,'' in \emph{IEEE Conference on Computer Vision and Pattern Recognition}, 2017, pp. 840--849.

\bibitem{bengio13}
Y.~Bengio, A.~Courville, and P.~Vincent, ``Representation learning: A review and new perspectives,'' \emph{IEEE Transactions on Pattern Analysis and Machine Intelligence}, vol.~35, no.~8, pp. 1798--1828, 2013.

\bibitem{Welsh02}
T.~Welsh, M.~Ashikhmin, and K.~Mueller, ``Transferring color to greyscale images,'' \emph{ACM Transactions on Graphics}, vol.~21, no.~3, pp. 277--280, 2002.

\bibitem{analogy}
A.~Hertzmann, C.~E. Jacobs, N.~Oliver, B.~Curless, and D.~H. Salesin, ``Image analogies,'' in \emph{ACM SIGGRAPH}, 2001, pp. 327--340.

\bibitem{Irony05}
R.~Irony, D.~Cohen-Or, and D.~Lischinski, ``Colorization by example,'' in \emph{Eurographics Symposium on Rendering}, 2005, pp. 201--210.

\bibitem{charpiat2008automatic}
G.~Charpiat, M.~Hofmann, and B.~Schölkopf, ``Automatic image colorization via multimodal predictions,'' in \emph{European Conference on Computer Vision}, 2008, pp. 126--139.

\bibitem{liu-2008-intrinsic}
X.~Liu, L.~Wan, Y.~Qu, T.-T. Wong, S.~Lin, C.-S. Leung, and P.-A. Heng, ``Intrinsic colorization,'' \emph{ACM Transactions on Graphics}, vol.~27, no.~5, pp. 152:1--152:9, 2008.

\bibitem{morimoto09}
Y.~Morimoto, Y.~Taguchi, and T.~Naemura, ``Automatic colorization of grayscale images using multiple images on the web,'' in \emph{ACM SIGGRAPH: Talks}, 2009, p. 59:1.

\bibitem{chia11}
A.~Y.-S. Chia, S.~Zhuo, R.~K. Gupta, Y.-W. Tai, S.-Y. Cho, P.~Tan, and S.~Lin, ``Semantic colorization with internet images,'' \emph{ACM Transactions on Graphics}, vol.~30, no.~6, pp. 1--8, 2011.

\bibitem{gupta2012image}
R.~K. Gupta, A.~Y.-S. Chia, D.~Rajan, E.~S. Ng, and Z.~Huang, ``Image colorization using similar images,'' in \emph{ACM International Conference on Multimedia}, 2012, pp. 369--378.

\bibitem{bugeau2013variational}
A.~Bugeau, V.-T. Ta, and N.~Papadakis, ``Variational exemplar-based image colorization,'' \emph{IEEE Transactions on Image Processing}, vol.~23, no.~1, pp. 298--307, 2014.

\bibitem{li17}
B.~Li, F.~Zhao, Z.~Su, X.~Liang, Y.-K. Lai, and P.~L. Rosin, ``Example-based image colorization using locality consistent sparse representation,'' \emph{IEEE Transactions on Image Processing}, vol.~26, no.~11, pp. 5188--5202, 2017.

\bibitem{he18}
M.~He, D.~Chen, J.~Liao, P.~V. Sander, and L.~Yuan, ``Deep exemplar-based colorization,'' \emph{ACM Transactions on Graphics}, vol.~37, no.~4, pp. 47:1--47:16, 2018.

\bibitem{li19}
B.~Li, Y.-K. Lai, M.~John, and P.~L. Rosin, ``Automatic example-based image colorization using location-aware cross-scale matching,'' \emph{IEEE Transactions on Image Processing}, vol.~28, no.~9, pp. 4606--4619, 2019.

\bibitem{xiao2020example}
C.~Xiao, C.~Han, Z.~Zhang, J.~Qin, T.-T. Wong, G.~Han, and S.~He, ``Example-based colourization via dense encoding pyramids,'' \emph{Computer Graphics Forum}, vol.~39, no.~1, pp. 20--33, 2020.

\bibitem{xu20}
Z.~Xu, T.~Wang, F.~Fang, Y.~Sheng, and G.~Zhang, ``Stylization-based architecture for fast deep exemplar colorization,'' in \emph{IEEE Conference on Computer Vision and Pattern Recognition}, 2020, pp. 9360--9369.

\bibitem{huang2017arbitrary}
X.~Huang and S.~Belongie, ``Arbitrary style transfer in real-time with adaptive instance normalization,'' in \emph{IEEE International Conference on Computer Vision}, 2017, pp. 1510--1519.

\bibitem{li2018closed}
Y.~Li, M.-Y. Liu, X.~Li, M.-H. Yang, and J.~Kautz, ``A closed-form solution to photorealistic image stylization,'' in \emph{European Conference on Computer Vision}, 2018, pp. 468--483.

\bibitem{lu20}
P.~Lu, J.~Yu, X.~Peng, Z.~Zhao, and X.~Wang, ``{Gray2ColorNet}: Transfer more colors from reference image,'' in \emph{ACM International Conference on Multimedia}, 2020, pp. 3210--3218.

\bibitem{fang20}
F.~Fang, T.~Wang, T.~Zeng, and G.~Zhang, ``A superpixel-based variational model for image colorization,'' \emph{IEEE Transactions on Visualization and Computer Graphics}, vol.~26, no.~10, pp. 2931--2943, 2020.

\bibitem{yin21}
W.~Yin, P.~Lu, Z.~Zhao, and X.~Peng, ``Yes, "attention is all you need", for exemplar based colorization,'' in \emph{ACM International Conference on Multimedia}, 2021, pp. 2243--2251.

\bibitem{li21}
H.~Li, B.~Sheng, P.~Li, R.~Ali, and C.~L.~P. Chen, ``Globally and locally semantic colorization via exemplar-based {Broad-GAN},'' \emph{IEEE Transactions on Image Processing}, vol.~30, pp. 8526--8539, 2021.

\bibitem{chen18}
C.~L.~P. Chen and Z.~Liu, ``Broad learning system: An effective and efficient incremental learning system without the need for deep architecture,'' \emph{IEEE Transactions on Neural Networks and Learning Systems}, vol.~29, no.~1, pp. 10--24, 2018.

\bibitem{Carrillo_2022_ACCV}
H.~Carrillo, M.~Clément, and A.~Bugeau, ``Super-attention for exemplar-based image colorization,'' in \emph{Asian Conference on Computer Vision}, 2022, pp. 646--662.

\bibitem{bai2022semantic}
Y.~Bai, C.~Dong, Z.~Chai, A.~Wang, Z.~Xu, and C.~Yuan, ``Semantic-sparse colorization network for deep exemplar-based colorization,'' in \emph{European Conference on Computer Vision}, 2022, pp. 505--521.

\bibitem{wang23}
H.~Wang, D.~Zhai, X.~Liu, J.~Jiang, and W.~Gao, ``Unsupervised deep exemplar colorization via pyramid dual non-local attention,'' \emph{IEEE Transactions on Image Processing}, vol.~32, pp. 4114--4127, 2023.

\bibitem{xu23}
R.~Xu, Z.~Tu, Y.~Du, X.~Dong, J.~Li, Z.~Meng, J.~Ma, A.~Bovik, and H.~Yu, ``{Pik-Fix:} restoring and colorizing old photos,'' in \emph{IEEE Winter Conference on Applications of Computer Vision}, 2023, pp. 1724--1734.

\bibitem{zhang17}
L.~Zhang, Y.~Ji, X.~Lin, and C.~Liu, ``Style transfer for anime sketches with enhanced residual u-net and auxiliary classifier {GAN},'' in \emph{IAPR Asian Conference on Pattern Recognition}, 2017, pp. 506--511.

\bibitem{akita2020colorization}
K.~Akita, Y.~Morimoto, and R.~Tsuruno, ``Colorization of line drawings with empty pupils,'' \emph{Computer Graphics Forum}, vol.~39, no.~7, pp. 601--610, 2020.

\bibitem{gatys16}
L.~A. Gatys, A.~S. Ecker, and M.~Bethge, ``Image style transfer using convolutional neural networks,'' in \emph{IEEE Conference on Computer Vision and Pattern Recognition}, 2016, pp. 2414--2423.

\bibitem{pmlr-v70-odena17a}
A.~Odena, C.~Olah, and J.~Shlens, ``Conditional image synthesis with auxiliary classifier {GAN}s,'' in \emph{International Conference on Machine Learning}, vol.~70, 2017, pp. 2642--2651.

\bibitem{fu17}
C.~Furusawa, K.~Hiroshiba, K.~Ogaki, and Y.~Odagiri, ``Comicolorization: Semi-automatic manga colorization,'' in \emph{ACM SIGGRAPH Asia Technical Briefs}, 2017, pp. 12:1--12:4.

\bibitem{lee20}
J.~Lee, E.~Kim, Y.~Lee, D.~Kim, J.~Chang, and J.~Choo, ``Reference-based sketch image colorization using augmented-self reference and dense semantic correspondence,'' in \emph{IEEE Conference on Computer Vision and Pattern Recognition}, 2020, pp. 5800--5809.

\bibitem{li2022eliminating}
Z.~Li, Z.~Geng, Z.~Kang, W.~Chen, and Y.~Yang, ``Eliminating gradient conflict in reference-based line-art colorization,'' in \emph{European Conference on Computer Vision}, 2022, pp. 579--596.

\bibitem{liu2022reference}
X.~Liu, W.~Wu, C.~Li, Y.~Li, and H.~Wu, ``Reference-guided structure-aware deep sketch colorization for cartoons,'' \emph{Computational Visual Media}, vol.~8, no.~1, pp. 135--148, 2022.

\bibitem{chen22}
S.-Y. Chen, J.-Q. Zhang, L.~Gao, Y.~He, S.~Xia, M.~Shi, and F.-L. Zhang, ``Active colorization for cartoon line drawings,'' \emph{IEEE Transactions on Visualization and Computer Graphics}, vol.~28, no.~2, pp. 1198--1208, 2022.

\bibitem{cao23}
Y.~Cao, H.~Tian, and P.~Y. Mok, ``Attention-aware anime line drawing colorization,'' in \emph{IEEE International Conference on Multimedia and Expo}, 2023, pp. 1637--1642.

\bibitem{wu23}
S.~Wu, X.~Yan, W.~Liu, S.~Xu, and S.~Zhang, ``Self-driven dual-path learning for reference-based line art colorization under limited data,'' \emph{IEEE Transactions on Circuits and Systems for Video Technology}, pp. 1--15, 2023.

\bibitem{animediffusion24}
Y.~Cao, X.~Meng, P.~Y. Mok, T.-Y. Lee, X.~Liu, and P.~Li, ``{AnimeDiffusion}: Anime diffusion colorization,'' \emph{IEEE Transactions on Visualization and Computer Graphics}, vol.~30, no.~10, pp. 6956--6969, 2024.

\bibitem{levin04}
A.~Levin, D.~Lischinski, and Y.~Weiss, ``Colorization using optimization,'' \emph{ACM Transactions on Graphics}, vol.~23, no.~3, pp. 689--694, 2004.

\bibitem{huang05}
Y.-C. Huang, Y.-S. Tung, J.-C. Chen, S.-W. Wang, and J.-L. Wu, ``An adaptive edge detection based colorization algorithm and its applications,'' in \emph{ACM International Conference on Multimedia}, 2005, pp. 351--354.

\bibitem{ys06}
L.~Yatziv and G.~Sapiro, ``Fast image and video colorization using chrominance blending,'' \emph{IEEE Transactions on Image Processing}, vol.~15, no.~5, pp. 1120--1129, 2006.

\bibitem{luan07}
Q.~Luan, F.~Wen, D.~Cohen-Or, L.~Liang, Y.-Q. Xu, and H.-Y. Shum, ``Natural image colorization,'' in \emph{Eurographics Symposium on Rendering}, 2007, pp. 309--320.

\bibitem{richard17}
R.~Zhang, J.-Y. Zhu, P.~Isola, X.~Geng, A.~S. Lin, T.~Yu, and A.~A. Efros, ``Real-time user-guided image colorization with learned deep priors,'' \emph{ACM Transactions on Graphics}, vol.~36, no.~4, pp. 119:1--119:11, 2017.

\bibitem{krizhevsky2012imagenet}
A.~Krizhevsky, I.~Sutskever, and G.~E. Hinton, ``{ImageNet} classification with deep convolutional neural networks,'' in \emph{Advances in Neural Information Processing Systems}, vol.~25, 2012, pp. 1097--1105.

\bibitem{kim21}
E.~Kim, S.~Lee, J.~Park, S.~Choi, C.~Seo, and J.~Choo, ``Deep edge-aware interactive colorization against color-bleeding effects,'' in \emph{IEEE International Conference on Computer Vision}, 2021, pp. 14\,647--14\,656.

\bibitem{xiao22}
Y.~Xiao, J.~Wu, J.~Zhang, P.~Zhou, Y.~Zheng, C.-S. Leung, and L.~Kavan, ``Interactive deep colorization and its application for image compression,'' \emph{IEEE Transactions on Visualization and Computer Graphics}, vol.~28, no.~3, pp. 1557--1572, 2022.

\bibitem{yun23}
J.~Yun, S.~Lee, M.~Park, and J.~Choo, ``{iColoriT}: Towards propagating local hints to the right region in interactive colorization by leveraging vision transformer,'' in \emph{IEEE Winter Conference on Applications of Computer Vision}, 2023, pp. 1787--1796.

\bibitem{dosovitskiy2020vit}
A.~Dosovitskiy, L.~Beyer, A.~Kolesnikov, D.~Weissenborn, X.~Zhai, T.~Unterthiner, M.~Dehghani, M.~Minderer, G.~Heigold, S.~Gelly, J.~Uszkoreit, and N.~Houlsby, ``An image is worth 16x16 words: Transformers for image recognition at scale,'' in \emph{International Conference on Learning Representations}, 2021, pp. 1--21.

\bibitem{qu06}
Y.~Qu, T.-T. Wong, and P.-A. Heng, ``Manga colorization,'' \emph{ACM Transactions on Graphics}, vol.~25, no.~3, pp. 1214--1220, 2006.

\bibitem{sykora2009lazybrush}
D.~Sýkora, J.~Dingliana, and S.~Collins, ``Lazybrush: Flexible painting tool for hand-drawn cartoons,'' \emph{Computer Graphics Forum}, vol.~28, no.~2, pp. 599--608, 2009.

\bibitem{sangkloy2017scribbler}
P.~Sangkloy, J.~Lu, C.~Fang, F.~Yu, and J.~Hays, ``Scribbler: Controlling deep image synthesis with sketch and color,'' in \emph{IEEE Conference on Computer Vision and Pattern Recognition}, 2017, pp. 6836--6845.

\bibitem{LIU201878}
Y.~Liu, Z.~Qin, T.~Wan, and Z.~Luo, ``Auto-painter: Cartoon image generation from sketch by using conditional wasserstein generative adversarial networks,'' \emph{Neurocomputing}, vol. 311, pp. 78--87, 2018.

\bibitem{mirza2014conditional}
M.~Mirza and S.~Osindero, ``Conditional generative adversarial nets,'' \emph{arXiv preprint arXiv:1411.1784}, pp. 1--7, 2014.

\bibitem{ci18}
Y.~Ci, X.~Ma, Z.~Wang, H.~Li, and Z.~Luo, ``User-guided deep anime line art colorization with conditional adversarial networks,'' in \emph{ACM International Conference on Multimedia}, 2018, pp. 1536--1544.

\bibitem{wgan-gp}
I.~Gulrajani, F.~Ahmed, M.~Arjovsky, V.~Dumoulin, and A.~C. Courville, ``Improved training of wasserstein {GANs},'' in \emph{Advances in Neural Information Processing Systems}, vol.~30, 2017, pp. 5767--5777.

\bibitem{johnson2016perceptual}
J.~Johnson, A.~Alahi, and F.-F. Li, ``Perceptual losses for real-time style transfer and super-resolution,'' in \emph{European Conference on Computer Vision}, 2016, pp. 694--711.

\bibitem{zhang18}
L.~Zhang, C.~Li, T.-T. Wong, Y.~Ji, and C.~Liu, ``Two-stage sketch colorization,'' \emph{ACM Transactions on Graphics}, vol.~37, no.~6, pp. 261:1--261:14, 2018.

\bibitem{zhang2021user}
L.~Zhang, C.~Li, E.~Simo-Serra, Y.~Ji, T.-T. Wong, and C.~Liu, ``User-guided line art flat filling with split filling mechanism,'' in \emph{IEEE Conference on Computer Vision and Pattern Recognition}, 2021, pp. 9884--9893.

\bibitem{yuan2021line}
M.~Yuan and E.~Simo-Serra, ``Line art colorization with concatenated spatial attention,'' in \emph{IEEE Conference on Computer Vision and Pattern Recognition Workshops}, 2021, pp. 3941--3945.

\bibitem{dou21}
Z.~Dou, N.~Wang, B.~Li, Z.~Wang, H.~Li, and B.~Liu, ``Dual color space guided sketch colorization,'' \emph{IEEE Transactions on Image Processing}, vol.~30, pp. 7292--7304, 2021.

\bibitem{carrillo2023diffusart}
H.~Carrillo, M.~Clément, A.~Bugeau, and E.~Simo-Serra, ``Diffusart: Enhancing line art colorization with conditional diffusion models,'' in \emph{IEEE Conference on Computer Vision and Pattern Recognition Workshops}, 2023, pp. 3486--3490.

\bibitem{cho23}
Y.~Cho, J.~Lee, S.~Yang, J.~Kim, Y.~Park, H.~Lee, M.~A. Khan, D.~Kim, and J.~Choo, ``Guiding users to where to give color hints for efficient interactive sketch colorization via unsupervised region prioritization,'' in \emph{IEEE Winter Conference on Applications of Computer Vision}, 2023, pp. 1818--1827.

\bibitem{wang2012affective}
X.-H. Wang, J.~Jia, H.-Y. Liao, and L.-H. Cai, ``Affective image colorization,'' \emph{Journal of Computer Science and Technology}, vol.~27, no.~6, pp. 1119--1128, 2012.

\bibitem{chang2015palette}
H.~Chang, O.~Fried, Y.~Liu, S.~DiVerdi, and A.~Finkelstein, ``Palette-based photo recoloring,'' \emph{ACM Transactions on Graphics}, vol.~34, no.~4, pp. 139:1--139:11, 2015.

\bibitem{Guadarrama17}
S.~Guadarrama, R.~Dahl, D.~Bieber, M.~Norouzi, J.~Shlens, and K.~Murphy, ``{PixColor:} pixel recursive colorization,'' in \emph{British Machine Vision Conference}, 2017, pp. 1--17.

\bibitem{royer2017probabilistic}
A.~Royer, A.~Kolesnikov, and C.~H. Lampert, ``Probabilistic image colorization,'' in \emph{British Machine Vision Conference}, 2017, pp. 1--15.

\bibitem{baig2017multiple}
M.~H. Baig and L.~Torresani, ``Multiple hypothesis colorization and its application to image compression,'' \emph{Computer Vision and Image Understanding}, vol. 164, pp. 111--123, 2017.

\bibitem{Deshpande17}
A.~Deshpande, J.~Lu, M.-C. Yeh, M.~J. Chong, and D.~Forsyth, ``Learning diverse image colorization,'' in \emph{IEEE Conference on Computer Vision and Pattern Recognition}, 2017, pp. 2877--2885.

\bibitem{cao17}
Y.~Cao, Z.~Zhou, W.~Zhang, and Y.~Yu, ``Unsupervised diverse colorization via generative adversarial networks,'' in \emph{Machine Learning and Knowledge Discovery in Databases}, 2017, pp. 151--166.

\bibitem{messaoud18}
S.~Messaoud, D.~Forsyth, and A.~G. Schwing, ``Structural consistency and controllability for diverse colorization,'' in \emph{European Conference on Computer Vision}, 2018, pp. 603--619.

\bibitem{wu21}
Y.~Wu, X.~Wang, Y.~Li, H.~Zhang, X.~Zhao, and Y.~Shan, ``Towards vivid and diverse image colorization with generative color prior,'' in \emph{IEEE International Conference on Computer Vision}, 2021, pp. 14\,357--14\,366.

\bibitem{cheng2015deep}
Z.~Cheng, Q.~Yang, and B.~Sheng, ``Deep colorization,'' in \emph{IEEE International Conference on Computer Vision}, 2015, pp. 415--423.

\bibitem{deshpande2015learning}
A.~Deshpande, J.~Rock, and D.~Forsyth, ``Learning large-scale automatic image colorization,'' in \emph{IEEE International Conference on Computer Vision}, 2015, pp. 567--575.

\bibitem{varga2017automatic}
D.~Varga, C.~A. Szabó, and T.~Szirányi, ``Automatic cartoon colorization based on convolutional neural network,'' in \emph{International Workshop on Content-Based Multimedia Indexing}, 2017, pp. 28:1--28:6.

\bibitem{dong2019learning}
X.~Dong, W.~Li, X.~Wang, and Y.~Wang, ``Learning a deep convolutional network for colorization in monochrome-color dual-lens system,'' in \emph{AAAI Conference on Artificial Intelligence}, vol.~33, no.~01, 2019, pp. 8255--8262.

\bibitem{dong2020colorization}
X.~Dong, W.~Li, X.~Hu, X.~Wang, and Y.~Wang, ``A colorization framework for monochrome-color dual-lens systems using a deep convolutional network,'' \emph{IEEE Transactions on Visualization and Computer Graphics}, vol.~28, no.~3, pp. 1469--1485, 2020.

\bibitem{yoo2019coloring}
S.~Yoo, H.~Bahng, S.~Chung, J.~Lee, J.~Chang, and J.~Choo, ``Coloring with limited data: Few-shot colorization via memory augmented networks,'' in \emph{IEEE Conference on Computer Vision and Pattern Recognition}, 2019, pp. 11\,275--11\,284.

\bibitem{alghofaili2021exploring}
R.~Alghofaili, M.~Fisher, R.~Zhang, M.~Lukáč, and L.-F. Yu, ``Exploring sketch-based character design guided by automatic colorization,'' in \emph{Graphics Interface}, 2021, pp. 56--67.

\bibitem{zhao2021scgan}
Y.~Zhao, L.-M. Po, K.-W. Cheung, W.-Y. Yu, and Y.~A.~U. Rehman, ``{SCGAN}: Saliency map-guided colorization with generative adversarial network,'' \emph{IEEE Transactions on Circuits and Systems for Video Technology}, vol.~31, no.~8, pp. 3062--3077, 2021.

\bibitem{jin2021focusing}
X.~Jin, Z.~Li, K.~Liu, D.~Zou, X.~Li, X.~Zhu, Z.~Zhou, Q.~Sun, and Q.~Liu, ``{Focusing on Persons}: Colorizing old images learning from modern historical movies,'' in \emph{ACM International Conference on Multimedia}, 2021, p. 1176–1184.

\bibitem{kim2022bigcolor}
G.~Kim, K.~Kang, S.~Kim, H.~Lee, S.~Kim, J.~Kim, S.-H. Baek, and S.~Cho, ``{BigColor}: Colorization using a generative color prior for natural images,'' in \emph{European Conference on Computer Vision}, 2022, pp. 350--366.

\bibitem{lee2022bridging}
H.~Lee, D.~Kim, D.~Lee, J.~Kim, and J.~Lee, ``Bridging the domain gap towards generalization in automatic colorization,'' in \emph{European Conference on Computer Vision}, 2022, pp. 527--543.

\bibitem{wang2022palgan}
Y.~Wang, M.~Xia, L.~Qi, J.~Shao, and Y.~Qiao, ``{PalGAN}: Image colorization with palette generative adversarial networks,'' in \emph{European Conference on Computer Vision}, 2022, pp. 271--288.

\bibitem{ardizzone2019guided}
L.~Ardizzone, C.~Lüth, J.~Kruse, C.~Rother, and U.~Köthe, ``Guided image generation with conditional invertible neural networks,'' \emph{arXiv preprint arXiv:1907.02392}, pp. 1--11, 2019.

\bibitem{kumar2021colorization}
M.~Kumar, D.~Weissenborn, and N.~Kalchbrenner, ``Colorization transformer,'' in \emph{International Conference on Learning Representations}, 2021, pp. 1--24.

\bibitem{ji2022colorformer}
X.~Ji, B.~Jiang, D.~Luo, G.~Tao, W.~Chu, Z.~Xie, C.~Wang, and Y.~Tai, ``{ColorFormer}: Image colorization via color memory assisted hybrid-attention transformer,'' in \emph{European Conference on Computer Vision}, 2022, pp. 20--36.

\bibitem{weng2022ct}
S.~Weng, J.~Sun, Y.~Li, S.~Li, and B.~Shi, ``{$CT^2$}: Colorization transformer via color tokens,'' in \emph{European Conference on Computer Vision}, 2022, pp. 1--16.

\bibitem{huang2022unicolor}
Z.~Huang, N.~Zhao, and J.~Liao, ``{UniColor}: A unified framework for multi-modal colorization with transformer,'' \emph{ACM Transactions on Graphics}, vol.~41, no.~6, pp. 205:1--205:16, 2022.

\bibitem{radford2021learning}
A.~Radford, J.~W. Kim, C.~Hallacy, A.~Ramesh, G.~Goh, S.~Agarwal, G.~Sastry, A.~Askell, P.~Mishkin, J.~Clark, G.~Krueger, and I.~Sutskever, ``Learning transferable visual models from natural language supervision,'' in \emph{International Conference on Machine Learning}, vol. 139, 2021, pp. 8748--8763.

\bibitem{saharia2022palette}
C.~Saharia, W.~Chan, H.~Chang, C.~Lee, J.~Ho, T.~Salimans, D.~Fleet, and M.~Norouzi, ``Palette: Image-to-image diffusion models,'' in \emph{ACM SIGGRAPH}, 2022, pp. 15:1--15:10.

\bibitem{liu2023improved}
H.~Liu, J.~Xing, M.~Xie, C.~Li, and T.-T. Wong, ``Improved diffusion-based image colorization via piggybacked models,'' \emph{arXiv preprint arXiv:2304.11105}, pp. 1--17, 2023.

\bibitem{zhang2023adding}
L.~Zhang, A.~Rao, and M.~Agrawala, ``Adding conditional control to text-to-image diffusion models,'' in \emph{IEEE International Conference on Computer Vision}, 2023, pp. 3836--3847.

\bibitem{zabari23}
N.~Zabari, A.~Azulay, A.~Gorkor, T.~Halperin, and O.~Fried, ``Diffusing colors: Image colorization with text guided diffusion,'' in \emph{ACM SIGGRAPH Asia}, 2023, pp. 61:1--61:11.

\bibitem{yatziv2006fast}
L.~Yatziv and G.~Sapiro, ``Fast image and video colorization using chrominance blending,'' \emph{IEEE Transactions on Image Processing}, vol.~15, no.~5, pp. 1120--1129, 2006.

\bibitem{jacob2009colorization}
V.~G. Jacob and S.~Gupta, ``Colorization of grayscale images and videos using a semiautomatic approach,'' in \emph{IEEE International Conference on Image Processing}, 2009, pp. 1653--1656.

\bibitem{sheng2013video}
B.~Sheng, H.~Sun, M.~Magnor, and P.~Li, ``Video colorization using parallel optimization in feature space,'' \emph{IEEE Transactions on Circuits and Systems for Video Technology}, vol.~24, no.~3, pp. 407--417, 2014.

\bibitem{ben2015approximate}
N.~Ben-Zrihem and L.~Zelnik-Manor, ``Approximate nearest neighbor fields in video,'' in \emph{IEEE Conference on Computer Vision and Pattern Recognition}, 2015, pp. 5233--5242.

\bibitem{bonneel2015blind}
N.~Bonneel, J.~Tompkin, K.~Sunkavalli, D.~Sun, S.~Paris, and H.~Pfister, ``Blind video temporal consistency,'' \emph{ACM Transactions on Graphics}, vol.~34, no.~6, pp. 196:1--196:9, 2015.

\bibitem{xia2016robust}
S.~Xia, J.~Liu, Y.~Fang, W.~Yang, and Z.~Guo, ``Robust and automatic video colorization via multiframe reordering refinement,'' in \emph{IEEE International Conference on Image Processing}, 2016, pp. 4017--4021.

\bibitem{paul2016spatiotemporal}
S.~Paul, S.~Bhattacharya, and S.~Gupta, ``Spatiotemporal colorization of video using 3d steerable pyramids,'' \emph{IEEE Transactions on Circuits and Systems for Video Technology}, vol.~27, no.~8, pp. 1605--1619, 2016.

\bibitem{jampani2017video}
V.~Jampani, R.~Gadde, and P.~V. Gehler, ``Video propagation networks,'' in \emph{IEEE Conference on Computer Vision and Pattern Recognition}, 2017, pp. 3154--3164.

\bibitem{meyer2018deep}
S.~Meyer, V.~Cornillère, A.~Djelouah, C.~Schroers, and M.~Gross, ``Deep video color propagation,'' in \emph{British Machine Vision Conference}, 2018, pp. 1--14.

\bibitem{lai2018learning}
W.-S. Lai, J.-B. Huang, O.~Wang, E.~Shechtman, E.~Yumer, and M.-H. Yang, ``Learning blind video temporal consistency,'' in \emph{European Conference on Computer Vision}, 2018, pp. 179--195.

\bibitem{liu2018switchable}
S.~Liu, G.~Zhong, S.~De~Mello, J.~Gu, V.~Jampani, M.-H. Yang, and J.~Kautz, ``Switchable temporal propagation network,'' in \emph{European Conference on Computer Vision}, 2018, pp. 89--104.

\bibitem{vondrick2018tracking}
C.~Vondrick, A.~Shrivastava, A.~Fathi, S.~Guadarrama, and K.~Murphy, ``Tracking emerges by colorizing videos,'' in \emph{European Conference on Computer Vision}, 2018, pp. 402--419.

\bibitem{zhang2019deep}
B.~Zhang, M.~He, J.~Liao, P.~V. Sander, L.~Yuan, A.~Bermak, and D.~Chen, ``Deep exemplar-based video colorization,'' in \emph{IEEE Conference on Computer Vision and Pattern Recognition}, 2019, pp. 8044--8053.

\bibitem{lei2019fully}
C.~Lei and Q.~Chen, ``Fully automatic video colorization with self-regularization and diversity,'' in \emph{IEEE Conference on Computer Vision and Pattern Recognition}, 2019, pp. 3748--3756.

\bibitem{iizuka2019deepremaster}
S.~Iizuka and E.~Simo-Serra, ``Deepremaster: temporal source-reference attention networks for comprehensive video enhancement,'' \emph{ACM Transactions on Graphics}, vol.~38, no.~6, pp. 176:1--176:13, 2019.

\bibitem{lei2020blind}
C.~Lei, Y.~Xing, and Q.~Chen, ``Blind video temporal consistency via deep video prior,'' in \emph{Advances in Neural Information Processing Systems}, vol.~33, 2020, pp. 1083--1093.

\bibitem{casey2021animation}
E.~Casey, V.~Pérez, and Z.~Li, ``The animation transformer: Visual correspondence via segment matching,'' in \emph{IEEE International Conference on Computer Vision}, 2021, pp. 11\,303--11\,312.

\bibitem{zhang2021line}
Q.~Zhang, B.~Wang, W.~Wen, H.~Li, and J.~Liu, ``Line art correlation matching feature transfer network for automatic animation colorization,'' in \emph{IEEE Winter Conference on Applications of Computer Vision}, 2021, pp. 3871--3880.

\bibitem{yang2022bistnet}
Y.~Yang, Z.~Peng, X.~Du, Z.~Tao, J.~Tang, and J.~Pan, ``{BiSTNet:} semantic image prior guided bidirectional temporal feature fusion for deep exemplar-based video colorization,'' \emph{arXiv preprint arXiv:2212.02268}, pp. 1--9, 2022.

\bibitem{shi2023reference}
M.~Shi, J.-Q. Zhang, S.-Y. Chen, L.~Gao, Y.-K. Lai, and F.-L. Zhang, ``Reference-based deep line art video colorization,'' \emph{IEEE Transactions on Visualization and Computer Graphics}, vol.~29, no.~6, pp. 2965--2979, 2023.

\bibitem{zhao2023svcnet}
Y.~Zhao, L.-M. Po, K.~Liu, X.~Wang, W.-Y. Yu, P.~Xian, Y.~Zhang, and M.~Liu, ``{SVCNet:} scribble-based video colorization network with temporal aggregation,'' \emph{IEEE Transactions on Image Processing}, vol.~32, pp. 4443--4458, 2023.

\bibitem{liu2023video}
H.~Liu, M.~Xie, J.~Xing, C.~Li, and T.-T. Wong, ``Video colorization with pre-trained text-to-image diffusion models,'' \emph{arXiv preprint arXiv:2306.01732}, pp. 1--13, 2023.

\bibitem{liu2024temporally}
Y.~Liu, H.~Zhao, K.~C. Chan, X.~Wang, C.~C. Loy, Y.~Qiao, and C.~Dong, ``Temporally consistent video colorization with deep feature propagation and self-regularization learning,'' \emph{Computational Visual Media}, vol.~10, no.~2, pp. 375--395, 2024.

\bibitem{chen2024exemplar}
S.~Chen, X.~Li, X.~Zhang, M.~Wang, Y.~Zhang, J.~Han, and Y.~Zhang, ``Exemplar-based video colorization with long-term spatiotemporal dependency,'' \emph{Knowledge-Based Systems}, vol. 284, pp. 1--12, 2024.

\bibitem{anwar2020image}
S.~Anwar, M.~Tahir, C.~Li, A.~Mian, F.~S. Khan, and A.~W. Muzaffar, ``Image colorization: A survey and dataset,'' \emph{arXiv preprint arXiv:2008.10774}, pp. 1--20, 2020.

\bibitem{HUANG2022105006}
S.~Huang, X.~Jin, Q.~Jiang, and L.~Liu, ``Deep learning for image colorization: Current and future prospects,'' \emph{Engineering Applications of Artificial Intelligence}, vol. 114, no.~C, pp. 1--27, 2022.

\bibitem{CHEN202251}
S.-Y. Chen, J.-Q. Zhang, Y.-Y. Zhao, P.~L. Rosin, Y.-K. Lai, and L.~Gao, ``A review of image and video colorization: From analogies to deep learning,'' \emph{Visual Informatics}, vol.~6, no.~3, pp. 51--68, 2022.

\bibitem{imagenet09}
J.~Deng, W.~Dong, R.~Socher, L.-J. Li, K.~Li, and F.-F. Li, ``{ImageNet:} a large-scale hierarchical image database,'' in \emph{IEEE Conference on Computer Vision and Pattern Recognition}, 2009, pp. 248--255.

\bibitem{pascal2010}
M.~Everingham, L.~Van~Gool, C.~K.~I. Williams, J.~Winn, and A.~Zisserman, ``The pascal visual object classes (voc) challenge,'' \emph{International Journal of Computer Vision}, vol.~88, pp. 303--338, 2010.

\bibitem{coco14}
T.-Y. Lin, M.~Maire, S.~Belongie, J.~Hays, P.~Perona, D.~Ramanan, P.~Dollár, and C.~L. Zitnick, ``{Microsoft COCO:} common objects in context,'' in \emph{European Conference on Computer Vision}, 2014, pp. 740--755.

\bibitem{places205}
B.~Zhou, A.~Lapedriza, J.~Xiao, A.~Torralba, and A.~Oliva, ``Learning deep features for scene recognition using places database,'' in \emph{Advances in Neural Information Processing Systems}, vol.~27, 2014, pp. 487--495.

\bibitem{ade20k17}
B.~Zhou, H.~Zhao, X.~Puig, S.~Fidler, A.~Barriuso, and A.~Torralba, ``Scene parsing through ade20k dataset,'' in \emph{IEEE Conference on Computer Vision and Pattern Recognition}, 2017, pp. 5122--5130.

\bibitem{by2011}
V.~Bychkovsky, S.~Paris, E.~Chan, and F.~Durand, ``Learning photographic global tonal adjustment with a database of input / output image pairs,'' in \emph{IEEE Conference on Computer Vision and Pattern Recognition}, 2011, pp. 97--104.

\bibitem{zou18}
C.~Zou, Q.~Yu, R.~Du, H.~Mo, Y.-Z. Song, T.~Xiang, C.~Gao, B.~Chen, and H.~Zhang, ``{SketchyScene:} richly-annotated scene sketches,'' in \emph{European Conference on Computer Vision}, 2018, pp. 438--454.

\bibitem{cifar09}
A.~Krizhevsky and G.~Hinton, ``Learning multiple layers of features from tiny images,'' \emph{Master's thesis, Department of Computer Science, University of Toronto}, 2009.

\bibitem{lsun15}
F.~Yu, A.~Seff, Y.~Zhang, S.~Song, T.~Funkhouser, and J.~Xiao, ``{LSUN:} construction of a large-scale image dataset using deep learning with humans in the loop,'' \emph{arXiv preprint arXiv:1506.03365}, pp. 1--9, 2015.

\bibitem{manga109-17}
Y.~Matsui, K.~Ito, Y.~Aramaki, A.~Fujimoto, T.~Ogawa, T.~Yamasaki, and K.~Aizawa, ``Sketch-based manga retrieval using manga109 dataset,'' \emph{Multimedia Tools and Applications}, vol.~76, pp. 21\,811--21\,838, 2017.

\bibitem{danboorucharacter}
\BIBentryALTinterwordspacing
Y.~Wang, ``Danbooru 2018 anime character recognition dataset,'' 2018. [Online]. Available: \url{https://github.com/grapeot/Danbooru2018AnimeCharacterRecognitionDataset}
\BIBentrySTDinterwordspacing

\bibitem{ascp}
\BIBentryALTinterwordspacing
T.~Kim, ``Anime sketch colorization pair,'' 2019. [Online]. Available: \url{https://www.kaggle.com/datasets/ktaebum/anime-sketch-colorization-pair}
\BIBentrySTDinterwordspacing

\bibitem{videvo}
\BIBentryALTinterwordspacing
``Free stock video footage,'' 2024. [Online]. Available: \url{https://www.videvo.net/}
\BIBentrySTDinterwordspacing

\bibitem{davis16}
F.~Perazzi, J.~Pont-Tuset, B.~McWilliams, L.~Van~Gool, M.~Gross, and A.~Sorkine-Hornung, ``A benchmark dataset and evaluation methodology for video object segmentation,'' in \emph{IEEE Conference on Computer Vision and Pattern Recognition}, 2016, pp. 724--732.

\bibitem{kinetics17}
W.~Kay, J.~Carreira, K.~Simonyan, B.~Zhang, C.~Hillier, S.~Vijayanarasimhan, F.~Viola, T.~Green, T.~Back, P.~Natsev \emph{et~al.}, ``The kinetics human action video dataset,'' \emph{arXiv preprint arXiv:1705.06950}, pp. 1--22, 2017.

\bibitem{truckenbrod81}
J.~R. Truckenbrod, ``Effective use of color in computer graphics,'' in \emph{ACM SIGGRAPH}, vol.~15, no.~3, 1981, p. 83–90.

\bibitem{li21cartoon}
X.~Li, B.~Zhang, J.~Liao, and P.~V. Sander, ``Deep sketch-guided cartoon video inbetweening,'' \emph{IEEE Transactions on Visualization and Computer Graphics}, vol.~28, no.~8, pp. 2938--2952, 2022.

\bibitem{deng2017}
Y.~Deng, C.~C. Loy, and X.~Tang, ``{Image Aesthetic Assessment:} an experimental survey,'' \emph{IEEE Signal Processing Magazine}, vol.~34, no.~4, pp. 80--106, 2017.

\bibitem{wang2004}
Z.~Wang, A.~C. Bovik, H.~R. Sheikh, and E.~P. Simoncelli, ``Image quality assessment: from error visibility to structural similarity,'' \emph{IEEE Transactions on Image Processing}, vol.~13, no.~4, pp. 600--612, 2004.

\bibitem{laion}
\BIBentryALTinterwordspacing
``Laion-aesthtics,'' 2022. [Online]. Available: \url{https://laion.ai/blog/laion-aesthetics/}
\BIBentrySTDinterwordspacing

\bibitem{wang2023exploring}
J.~Wang, K.~C. Chan, and C.~C. Loy, ``Exploring clip for assessing the look and feel of images,'' in \emph{AAAI Conference on Artificial Intelligence}, vol.~37, no.~2, 2023, pp. 2555--2563.

\bibitem{wu2023}
S.~Wu, Y.~Yang, S.~Xu, W.~Liu, X.~Yan, and S.~Zhang, ``{FlexIcon:} flexible icon colorization via guided images and palettes,'' in \emph{ACM International Conference on Multimedia}, 2023, p. 8662–8673.

\bibitem{dai2024learning}
Y.~Dai, S.~Zhou, Q.~Li, C.~Li, and C.~C. Loy, ``Learning inclusion matching for animation paint bucket colorization,'' in \emph{IEEE Conference on Computer Vision and Pattern Recognition}, 2024, pp. 25\,544--25\,553.

\bibitem{lin24}
J.~Lin, W.~Zhao, and Y.~Wang, ``Visual correspondence learning and spatially attentive synthesis via transformer for exemplar-based anime line art colorization,'' \emph{IEEE Transactions on Multimedia}, vol.~26, pp. 6880--6890, 2024.

\bibitem{liang2024control}
Z.~Liang, Z.~Li, S.~Zhou, C.~Li, and C.~C. Loy, ``{Control Color:} multimodal diffusion-based interactive image colorization,'' \emph{arXiv preprint arXiv:2402.10855}, pp. 1--10, 2024.

\bibitem{sun19}
T.-H. Sun, C.-H. Lai, S.-K. Wong, and Y.-S. Wang, ``Adversarial colorization of icons based on contour and color conditions,'' in \emph{ACM International Conference on Multimedia}, 2019, p. 683–691.

\bibitem{li22}
Y.~Li, Y.~Lien, and Y.~Wang, ``Style-structure disentangled features and normalizing flows for diverse icon colorization,'' in \emph{IEEE Conference on Computer Vision and Pattern Recognition}, 2022, pp. 11\,234--11\,243.

\bibitem{du2024palette}
Z.-J. Du, J.-W. Zhou, Z.-X. Xia, B.-F. Seng, and K.~Xu, ``Palette-based content-aware image recoloring,'' in \emph{Computational Visual Media}, 2024, pp. 240--258.

\bibitem{cho17}
J.~Cho, S.~Yun, K.~Lee, and J.~Y. Choi, ``{PaletteNet:} image recolorization with given color palette,'' in \emph{IEEE Conference on Computer Vision and Pattern Recognition Workshops}, 2017, pp. 1058--1066.

\bibitem{yan2024colorizediffusion}
D.~Yan, L.~Yuan, Y.~Nishioka, I.~Fujishiro, and S.~Saito, ``{ColorizeDiffusion:} adjustable sketch colorization with reference image and text,'' \emph{arXiv preprint arXiv:2401.01456}, pp. 1--17, 2024.

\bibitem{xing2024tooncrafter}
J.~Xing, H.~Liu, M.~Xia, Y.~Zhang, X.~Wang, Y.~Shan, and T.-T. Wong, ``{ToonCrafter:} generative cartoon interpolation,'' \emph{arXiv preprint arXiv:2405.17933}, pp. 1--12, 2024.

\bibitem{duan2024}
X.~Duan, Y.~Cao, R.~Zhang, X.~Wang, and P.~Li, ``Shadow-aware image colorization,'' \emph{The Visual Computer}, vol.~40, no.~7, pp. 4969--4979, 2024.

\bibitem{yan23}
D.~Yan, R.~Ito, R.~Moriai, and S.~Saito, ``{Two-Step Training:} adjustable sketch colourization via reference image and text tag,'' \emph{Computer Graphics Forum}, vol.~42, no.~6, pp. 1--14, 2023.

\bibitem{huang2024lvcd}
Z.~Huang, M.~Zhang, and J.~Liao, ``{LVCD:} reference-based lineart video colorization with diffusion models,'' \emph{arXiv preprint arXiv:2409.12960}, pp. 1--11, 2024.

\bibitem{sketchKeras}
\BIBentryALTinterwordspacing
lllyasviel, ``sketchkeras,'' 2017. [Online]. Available: \url{https://github.com/lllyasviel/sketchKeras/}
\BIBentrySTDinterwordspacing

\bibitem{Anime2Sketch}
\BIBentryALTinterwordspacing
Mukosame, ``Anime2sketch,'' 2021. [Online]. Available: \url{https://github.com/Mukosame/Anime2Sketch/}
\BIBentrySTDinterwordspacing

\bibitem{ebsynth}
\BIBentryALTinterwordspacing
jamriska, ``ebsynth,'' 2018. [Online]. Available: \url{https://github.com/jamriska/ebsynth/}
\BIBentrySTDinterwordspacing

\bibitem{yang23}
S.~Yang, Y.~Zhou, Z.~Liu, and C.~C. Loy, ``{Rerender A Video:} zero-shot text-guided video-to-video translation,'' in \emph{ACM SIGGRAPH Asia}, 2023, pp. 95:1--95:11.

\bibitem{qi23}
C.~Qi, X.~Cun, Y.~Zhang, C.~Lei, X.~Wang, Y.~Shan, and Q.~Chen, ``Fatezero: Fusing attentions for zero-shot text-based video editing,'' in \emph{IEEE International Conference on Computer Vision}, 2023, pp. 15\,886--15\,896.

\bibitem{zhao21}
N.~Zhao, Q.~Zheng, J.~Liao, Y.~Cao, H.~Pfister, and R.~W.~H. Lau, ``Selective region-based photo color adjustment for graphic designs,'' \emph{ACM Transactions on Graphics}, vol.~40, no.~2, pp. 17:1--17:17, 2021.

\bibitem{ke23}
Z.~Ke, Y.~Liu, L.~Zhu, N.~Zhao, and R.~W. Lau, ``Neural preset for color style transfer,'' in \emph{IEEE Conference on Computer Vision and Pattern Recognition}, 2023, pp. 14\,173--14\,182.

\end{thebibliography}

\end{document}